\documentclass[12pt]{arXiv_submission} 

\makeatletter
\def\set@curr@file#1{\def\@curr@file{#1}} 
\makeatother
\usepackage[load-configurations=version-1]{siunitx}

\newcommand*{\bbC}{\mathbb{C}}

\newcommand*{\bbN}{\mathbb{N}}

\newcommand*{\bbR}{\mathbb{R}}

\newcommand*{\cA}{\mathcal{A}}
\newcommand*{\cB}{\mathcal{B}}

\newcommand*{\cD}{\mathcal{D}}
\newcommand*{\cE}{\mathcal{E}}

\newcommand*{\cH}{\mathcal{H}}

\newcommand*{\cJ}{\mathcal{J}}

\newcommand*{\cL}{\mathcal{L}}

\newcommand*{\cN}{\mathcal{N}}

\newcommand*{\cO}{\mathcal{O}}
\newcommand*{\cP}{\mathcal{P}}

\newcommand*{\cU}{\mathcal{U}}
\newcommand*{\cT}{\mathcal{T}}
\newcommand*{\cV}{\mathcal{V}}

\def\bnu{\vec{\nu}}

\def\brho{\vec{\rho}}

\def\bnu{\vec{\nu}}

\def\bA{\vec{A}}

\def\x{\vec{x}}
\def\y{\vec{y}}
\def\0{\vec{0}}

\def\z{\vec{z}}

\def\D{\,\mathrm{d}}
\def\E{\mathrm{e}}
\newcommand*{\ip}[2]{\langle #1, #2 \rangle}
\newcommand*{\bm}[1]{\vec{#1}}
\newcommand*{\nm}[1]{\|{#1}\|}
\newcommand*{\nmu}[1]{{\|#1\|}}
\newcommand\qan{{\quad\hbox{and}\quad}}

\DefineNamedColor{named}{Purple}{cmyk}{0.45,0.86,0,0}

\DefineNamedColor{named}{JungleGreen} {cmyk}{0.99,0,0.52,0}

\DefineNamedColor{named}{orange} {rgb}{1,0.55,0}

\DefineNamedColor{named}{forestgreen} {rgb}{0.13,0.54,0.13}

\DefineNamedColor{named}{oceanfoam} {rgb}{0.34, 0.52, 0.54}

\DefineNamedColor{named}{burgundy} {rgb}{0.61, 0, 0}

\newcommand{\be}[1]{\begin{equation} #1 \end{equation}}
\newcommand{\bes}[1]{\begin{equation*} #1 \end{equation*}}

\newcommand{\eas}[1]{\begin{align*} #1 \end{align*}}

\newcommand\numberthis{\addtocounter{equation}{1}\tag{\theequation}}

\usepackage{enumerate}

\newtheorem{problem}{Problem}

\title[DNNs for Learning High-Dimensional, Hilbert-Valued Functions]
{Deep Neural Networks Are Effective At Learning High-Dimensional Hilbert-Valued Functions From Limited Data}

\usepackage{times}
\arXivauthor{\Name{Ben Adcock} \Email{ben\_adcock@sfu.ca}\\
  \addr Department of Mathematics, Simon Fraser University, Canada
\AND
 \Name{Simone Brugiapaglia} \Email{simone.brugiapaglia@concordia.ca}\\
 \addr Department of Mathematics and Statistics, Concordia University
   \AND
  \Name{Nick Dexter} \Email{nicholas\_dexter@sfu.ca}\\
   \addr Department of Mathematics, Simon Fraser University, Canada
    \AND
  \Name{Sebastian Moraga} \Email{smoragas@sfu.ca}\\
  \addr Department of Mathematics, Simon Fraser University, Canada}

\begin{document}

\maketitle

\begin{abstract}
The accurate approximation of scalar-valued functions from sample points is a key task in mathematical modelling and computational science. Recently, machine learning techniques based on Deep Neural Networks (DNNs) have begun to emerge as promising tools for function approximation in scientific computing problems, with some impressive results achieved on problems where the dimension of the underlying data or problem domain is large. In this work, we broaden this perspective by focusing on the approximation of functions that are \textit{Hilbert-valued}, i.e.\ they take values in a separable, but typically infinite-dimensional, Hilbert space. This problem arises in many science and engineering problems, in particular those involving the solution of parametric Partial Differential Equations (PDEs). Such problems are challenging for three reasons. First, pointwise samples are expensive to acquire. Second, the domain of the function is usually high dimensional, and third, the range lies in a Hilbert space. Our contributions are twofold. First, we present a novel result on DNN training for holomorphic functions with so-called \textit{hidden anisotropy}. This result introduces a DNN training procedure and a full theoretical analysis with explicit guarantees on the error and sample complexity. This error bound is explicit in the three key errors occurred in the approximation procedure: the best approximation error, the measurement error and the physical discretization error. Our result shows that there exists a procedure (albeit a non-standard one) for learning Hilbert-valued functions via DNNs that performs as well as, but no better than current best-in-class schemes. It therefore gives a benchmark lower bound for how well methods DNN training can perform on such problems. Second, we examine whether better performance can be achieved in practice through different types of architectures and training. We provide preliminary numerical results illustrating the practical performance of DNNs on Hilbert-valued functions arising as solutions to parametric PDEs. We consider different parameters, modify the DNN architecture to achieve better and competitive results and compare these to current best-in-class schemes. 
\end{abstract}


\section{Introduction}

Driven by their success in many historically-challenging machine learning problems, Deep Neural Networks (DNNs) and Deep Learning (DL) are beginning to be applied successfully to challenging tasks in computational science and engineering. Such tasks are often characterized by the high dimensionality of their data or problem. Examples include inverse problems in imaging \cite{adcock2021compressive,ongie2020inverse}, molecular dynamics simulations \cite{faber2017molecular}, protein structure prediction \cite{alphafold2020}, discovery of unknown dynamical systems \cite{Lagergren2020}, Partial Differential Equations (PDEs) \cite{berg2018pdes} and, as discussed below, parameterized PDEs for Uncertainty Quantification (UQ).

The application of DL to such problems is supported by a rapidly-growing theory on the approximation properties of DNNs (see, e.g., \cite{Yarotsky2017,Bach2017,Petersen2018,Beck2019,Grohs2019}). This has become an extremely active area, with many new results produced within the last several years. Generalizing the classical universal approximation theorem \cite{Cybenko1989,HornikKurt1989Mfna,LeshnoEtAl1993}, recent works have shown approximation results for DNNs in terms of their depth \cite{Liang2017,Lu2020,Yarotsky2018}, for functions in Sobolev spaces \cite{Guhring2019}, H\"older spaces \cite{Shen2019} and Barron spaces \cite{e2019barron}, for bandlimited functions \cite{Montanelli2017} and holomorphic functions \cite{e2018exponential,opschoor2019exponential}, as well as for tasks in scientific computing such as approximation of high-dimensional functions \cite{Schwab2017,Li2019}  and PDEs \cite{Grohs2018,berner2020analysis}, dimensionality reduction \cite{Zhang2019}, and methods for DEs \cite{Lu2017,Weinan2018}. Other works have established connections between DNNs and classical methods of approximation such as polynomials \cite{Schwab2017,Daws2019b}, splines \cite{Unser2019}, sparse grids \cite{MontanelliEtAl2019} and finite elements \cite{opschoor2019exponential}.  

The above list represents only a selection of the many recent results in this area. However, it is notable that these results generally fall into the category of \textit{existence theory}: namely, they assert the existence of a DNN with desirable approximation properties, but not a constructive means to compute such a network (we note in passing several exceptions \cite{Dereventsov2019b,Fokina2019,Daws2019a}, although these generally lack theoretical guarantees on trainability).
As discussed in \cite{geist2020numerical} and shown in \cite{adcock2020gap}, there can often be a substantial gap between theoretical existence results and practical performance of DNNs when trained using standard tools. 

Motivated by such a performance gap, this paper considers the following three key issues:

\textbf{(1) } The vast majority of previous work considers only scalar-valued function approximation.

\textbf{(2)} Existence theory says little about whether such a DNN can be obtained by training and the number of samples of the function needed to do so.

\textbf{(3)} Many applications in computational science are relatively \textit{data starved}. Hence it is critical to understand the \textit{sample complexity} DNN approximation: namely, how much data is needed to train an accurate DNN for a given function.

Specifically, the focus of this paper is on learning \textit{high-dimensional}, \textit{Hilbert-valued} functions from \textit{limited datasets} using DNNs. We next describe the motivations for considering this problem.

\subsection{Motivations}

An important task in UQ involves constructing a surrogate model of a physical system that depends on a set of parameters $\bm{y} \in \bbR^d$. The physical system is typically modelled via a PDE (or system of PDEs) in terms of the relevant spatial and temporal variables.  In other words, its solution is a function $u = u(\bm{x},\bm{y})$, where $\bm{x} \in \bbR^n$, $n = 1, 2,3,4$, 
denotes the physical variables (space and time) and $\bm{y}$ denotes the parameters. The function $u$ is the solution to a PDE system
\be{
\label{pPDEprob}
\cD_{\bm{x}} (u,\bm{y}) = 0,
}
where $\cD_{\bm{x}}(\cdot,\bm{y})$ is a differential operator in the physical variable $\bm{x}$ that depends on the parametric variable $\bm{y}$. The objective in surrogate model construction is to understand the parametric dependence of the solution $u$. Since the PDE problem can often be posed (in weak form) in a separable Hilbert space $\cV$, this is equivalent to approximating the \textit{Hilbert-valued} function
\be{\label{soln_map}
\bm{y} \in \bbR^d \mapsto u(\bm{y}) \in \cV,
}
(we suppress the $\bm{x}$ dependence for ease of notation). This problem have several key features that motivate this work (see, e.g.\ \cite{cohen2015high,Gunzburger2014} for more details):

\textbf{(i)} The input dimension $d$ is typically high. Indeed, the more parameters, the better the model for the physical system. Typically, $d$ may range from ten to over a hundred. Moreover, in some situations, one may also consider functions with infinitely-many parameters, i.e.\ $d = \infty$.

\textbf{(ii)} The output $u(\bm{y})$ takes values in a Hilbert space $\cV$. In some applications one may only wish to approximate some scalar \textit{Quantity-of-Interest (QoI)} of $u$ (e.g.\ the spatial mean $f(\bm{y}) = \int_{\Omega} u(\bm{x},\bm{y}) \D \bm{x}$ over the physical domain $\Omega \subseteq \bbR^n$). However, other applications call for approximating the whole solution $u(\bm{y})$ (from which one can obviously approximate any number of QoI's).

\textbf{(iii)} Computing samples is expensive. For each value of $\bm{y}$, evaluating $u(\bm{y})$ requires solving the PDE \eqref{pPDEprob} via a computationally-intensive numerical simulation. In practice, generating the samples may take a time ranging from minutes to days or even weeks. Furthermore, since the PDE is never solved exactly, this process always commits an error -- the \textit{measurement error} as we term it.

To summarize, surrogate model construction involves approximating a high-dimensional, Hilbert-valued function from limited data. Such a task is clearly impossible without further assumptions. Fortunately, $u$ is often \textit{smooth}. To be precise, under certain conditions on the problem \eqref{pPDEprob} one can show that $u$ is a \textit{holomorphic} function of the parameters $\bm{y}$ \cite{cohen2010convergence, cohen2011analytic, chkifa2015breaking} (see also \cite{cohen2015high} for an in-depth review). This opens the door for approximating $u$ efficiently from limited samples, even when $d$ is large, and thereby lessening the \textit{curse of dimensionality}. We shall exploit this assumption throughout.

Even with this assumption, though, there is another hurdle to overcome; namely the infinite dimensionality of $\cV$. The usual way to address this is to introduce a finite-dimensional \textit{discretization} $\cV_h$, where $h$ is a discretization parameter, and compute an approximation taking values in $\cV_h$. Typically, since $u$ is the solution of a PDE, one takes $\cV_h$ to be a finite element discretization. However, regardless of how $\cV_h$ is chosen, it is important that the \textit{discretization error} (the effect of replacing $\cV$ by $\cV_h$) be quantified explicitly in the overall error bound. We also address this matter.

\subsection{Contributions}

In this work we study the approximation of a holomorphic, Hilbert-valued function $f : \cU \rightarrow \cV$
(we now switch notation from $u$ to $f$ since the problem we consider does not necessarily need to arise as the solution of a parametric PDE) from $m$ noisy sample values
\bes{
d_i = f(\bm{y}_i) + n_i, \quad \forall i = 1,\ldots,m.
}
We assume throughout that $\cU = [-1,1]^d$ is the unit hypercube 
 and the $\bm{y}_i$ are drawn identically and independently from the uniform measure on $\cU$. Note that this choice of sampling is not only typical in practice, it is critical in allowing for theoretical sample complexity estimates that scale efficiently with the dimension $d$. The choice of the hypercube implies that the parameters $y_1,\ldots,y_d$ (where $\bm{y} = (y_i)^d_{i=1}$) 
 are independent and vary in finite intervals (which, up to rescaling, can be taken to be equal to $[-1,1]$). Both assumptions are also standard in practice.

The values $n_i$ are the \textit{measurement errors}. They constitute the errors involved in computing $f$, whether it be via a numerical PDE solve or some other unspecified process. Throughout, we assume that $\cV$ is a separable Hilbert space and that it is discretized via a finite-dimensional subspace $\cV_h \subseteq \cV$. We make the additional assumption that the samples $d_i$ are elements of $\cV_h$, i.e.
\bes{
d_i \in \cV_h, \quad \forall i = 1,\ldots,m,
}
and we seek to compute an approximation $\tilde{f} : \cU \rightarrow \cV_h$ to $f$ taking values in $\cV_h$.

Our first main contribution is a novel theoretical result, Theorem \ref{t:mainthm1}. It shows that there exists a DNN architecture (of a given size and depth depending on $m$ and $d$) and  a training procedure (i.e.\ a loss function)
such that any minimizer of the corresponding loss function minimization problem approximates $f$ up to an explicit error bound. This error bound splits into three key terms that fully describe the effects of all main errors involved in the approximation process:

\textbf{(a)} an \textit{approximation error} that is exponentially-small in $m^{1/(2d)}$, up to log factors;

\textbf{(b)} a \textit{measurement error} that is proportional to the $\ell^2$-norm of the measurement noise $( n_i)^{m}_{i=1}$;

\textbf{(c)} a \textit{physical discretization error} that is proportional to the best approximation error of $f$ in $\cV_h$.
\\
We term this result a \textit{practical existence theorem} for Hilbert-valued function approximation. Going beyond standard existence theory, as discussed above, it shows not only the existence of a DNN with desirable approximation properties, but both a means of obtaining it via training and an estimate on the sample complexity via the exponentially-decaying approximation error (a).

A key facet of this work is the assumption on $f$.
While we assume $f$ is holomorphic, we do not assume any further information on it. In particular, $f$ may have \textit{anisotropic} dependence on the parameters $y_1,\ldots,y_d$ -- i.e.\ it may vary more rapidly in some directions than in others -- and such anisotropy may be \textit{hidden} -- i.e.\ it is not used to construct the approximation. This \textit{hidden anisotropy} assumption is highly relevant in practice, yet substantially harder to tackle than scenarios where such behaviour is known \textit{a priori}. We formalize exactly what we mean by hidden anisotropy in \S \ref{s:holomorphyandstuff}.

As we note below, current best-in-class methods for holomorphic, Hilbert-valued function approximation are based on multivariate orthogonal polynomials and also achieve exponential rates of convergence in $m^{1/(2d)}$. Through (a) we show that the obtained DNN achieves the same rate of convergence, up to a constant. Hence this work shows the existence of a DNN training procedure that can perform as well as current best-in-class methods.

Further, we also categorize all the errors in the DNN training procedure, including the measurement error (b) and physical discretization error (c). The latter is ubiquitous in simulations (due to the need to work with $\cV_h$ instead of $\cV$), but often not included in theoretical analyses. Our work thus sheds light on understanding how to tune the method parameters ($m$, $h$, and so forth) to balance the various errors, thus leading to optimal practical performance. It also raises the potential for the use of multilevel or multifidelity schemes, where $m$ and $h$ are simultaneously refined.

Note that we establish Theorem \ref{t:mainthm1} by carefully re-interpreting a polynomial-based approximation based on compressed sensing as a DNN training procedure in which all the weights and biases are fixed, besides those in the final layer. We do not claim that this training procedure is practically advantageous to use -- indeed, since it mimics a polynomial approximation by construction, it is unlikely to offer better performance than the latter. The purpose of Theorem \ref{t:mainthm1} is to show that there are provably good ways to setup and train DNNs for holomorphic function approximation, even if these are not standard procedures. It highlights the potential to achieve better performance in practice with trained DNNs, by modifying the training setup and DNN architectures suitably. Further, since polynomial-based methods are strongly tied to the underlying smoothness of the functions being approximated, while DNNs are not, it suggests DNNs can be useful flexible tools, with the potential of achieving good performance across a range of different function classes.

Nonetheless, with this gap between theory and practice in mind, we end this paper by presenting initial numerical experiments showing the effectiveness of trained DNNs for parametric PDE problems. We also compare with best-in-class compressed sensing-based polynomial approximation schemes, with a focus on the sample complexity of both approaches. Theorem \ref{t:mainthm1} shows that there exists a DNN architecture and training procedure that performs at least as well as such schemes in this regard, but, as noted, this setup is neither standard, nor expected to yield any better performance in practice.
Following \cite{adcock2020gap}, in our experiments we use standard DNN architectures coupled with standard loss functions and training algorithms. We show that with proper architecture and hyperparameter selection we are able to achieve competitive results. In particular, we can use smaller architectures than those suggested by Theorem \ref{t:mainthm1} and the standard $\ell^2$-loss functions, provided we train all the weights and biases of the DNN. Further, we show that we can actually outperform state-of-the-art polynomial methods for the problem considered by switching from the ReLU (as used in Theorem \ref{t:mainthm1}) to smoother activations functions. All in all, these preliminary results demonstrate the promise of the DNN approach for parametric PDEs in terms of sample complexity. This complements recent results shown in \cite{geist2020numerical}, which showed favourable performance of DNNs with respect to the dimension $d$.

\subsection{Related work}\label{ss:previous_work}

There are various ways to approximate the solution map of a parametric PDE, including \textit{reduced basis methods} \cite{hesthaven2015certified} and \textit{polynomial chaos expansions} \cite{xiu2002wiener}. In this paper, we focus on comparing DNN performance against the latter. Polynomial expansions are well-suited to smooth function approximation, with the so-called best $s$-term polynomial approximation offering exponential rates of convergence in $s^{1/d}$ for holomorphic functions (see \S \ref{s:holomorphyandstuff}). For a function with isotropic or known anisotropic behaviour in its variables, such an approximation can be computed in a number of ways, including interpolation or least squares \cite{cohen2018multivariate}. The situation becomes more challenging when the anisotropy is hidden. Adaptive interpolation or least-squares schemes \cite{chkifa2013sparse,chkifa2014high,cohen2018multivariate,gittelson2013adaptive,migliorati2015adaptive} may work in practice, though they generally lack theoretical guarantees. 

In the last five years, techniques based on compressed sensing have emerged as viable tools for this problem (see, e.g., \cite{adcock2017compressed,adcock2019correcting,chkifa2018polynomial,doostan2011nonadapted,hampton2015compressive} 
and references therein). Theoretical guarantees show that such techniques provide quasi-best $s$-term polynomial approximations, with favourable sample complexities \cite{adcock2018infinite,adcock2017compressed,chkifa2018polynomial}.
However, standard compressed sensing only allows for recovery of real or complex sparse vectors, e.g., scalar QoI's of the solution map \eqref{soln_map}, and therefore does not allow for recovery guarantees for Hilbert-valued functions. The {\em Simultaneous Compressed Sensing} (SCS) method \cite{dexter2019mixed} enables fully discrete approximation of the solution map by combining spatial discretization, e.g., finite elements, with joint-sparse vector recovery techniques modified for recovery in $\cV_h$.
The SCS framework also extends theoretical recovery guarantees from compressed sensing to the Hilbert-valued setting, thereby inheriting the same quasi-best $s$-term approximation rates and sample complexity estimates.
Because of its favourable behaviour and theoretical guarantees, we refer to SCS as the current best-in-class method, and seek to match (or beat) this performance with a DNN procedure.

Recently, a number of works have applied DNNs to parametric PDEs. See \cite{pmlr-v107-cyr20a,dalsanto2020data,geist2020numerical,khoo2020solving,laakmann2020efficient} and references therein. These works generally lack theoretical analysis. On the theoretical side, \cite{opschoor2019exponential} provides an existence theorem for parametric PDEs with holomorphic solution maps. The work \cite{kutyniok2020theoretical} considers the reduced basis approach to parametric PDEs, showing the existence of a DNN whose size depends on the intrinsic low-dimensionality of the solution manifold. Neither result addresses whether such a DNN can be trained, nor the sample complexity in doing so. Our work in particular complements \cite{opschoor2019exponential} by showing that DNNs admitting the same approximation error can be obtained as solutions of certain training problems. Furthermore, our result also makes all errors committed by the process explicit, including the aforementioned measurement and physical discretization errors, which have typically not been addressed in previous works.

\section{Learning Hilbert-valued functions via DNNs}\label{s:setup}

We first require some notation. Throughout $d \in \bbN$ denotes the dimension of the input space. We write $\mathbb{N}_0^d:= \lbrace \vec{\nu} =(\nu_ k)_{k=1}^d: \nu_k \in \mathbb{N}_0 \rbrace$ for the set of nonnegative integer multi-indices and $\vec{0}$ and $\vec{1}$ for the multi-indices consisting of all zeros and all ones respectively. The inequality $\vec{\mu} \leq \vec{\nu}$ between two multi-indices is understood componentwise.

For $1 \leq p \leq \infty$, we write $\nm{\cdot}_{p}$ for the usual vector $\ell^p$-norm and for the induced matrix $\ell^p$-norm. Moreover, for $1 \leq p,q < \infty$ we define the matrix $\ell^{p,q}$-norm as  $\| \bm{G}\|^q_{p,q} := \sum_{j=1}^n \left ( \sum^{m}_{i=1} | G_{ij} |^p \right )^{q/p}$.
We also use the notation $A \lesssim B$ to mean that there exists a numerical constant $c > 0$ independent of $A$ and $B$ such that $A \leq c B$, and likewise for $A \gtrsim B$.

\subsection{Setup}\label{ss:setup}
We now describe the setup in detail.
We write $\bm{y} = (y_1,\ldots,y_d)$ for the variables and let $\cU = [-1,1]^d$.
We consider the uniform probability measure on $\cU$, i.e.
\begin{equation}\label{meas-unif}
\D \varrho(\bm{y}) =  2^{-d} \D \bm{y},\quad \forall\bm{y} \in \cU,
\end{equation}
and write $L^p_{\varrho}(\cU)$ for the corresponding weighted Lebesgue spaces of scalar-valued functions over $\cU$ and $\nm{\cdot}_{L^p_{\varrho} (\cU)}$ for their norms.

Throughout, we let $\cV$ be a separable Hilbert space over the field $\bbR$ (we could also consider the field $\bbC$ with few additional difficulties), with inner product $\ip{\cdot}{\cdot}_{\cV}$ and corresponding norm
\[
\|{v}\|_{\cV} :=\sqrt{\ip{v}{v}_{\cV}}.
\]
We let $\cV^{N}$ be the vector space of Hilbert-valued vectors of length $N$, i.e.\ $\vec{\nu} = (\nu_i)^{N}_{i=1}$ where $\nu_i \in \cV$, $i=1,\ldots,N$. More generally, let $\Lambda \subseteq \bbN^d_0$ denote a (possibly infinite) multi-index set. We write $\vec{v} = (v_{\vec{\nu}})_{\vec{\nu} \in \Lambda}$ for a sequence with $\cV$-valued entries, $v_{\vec{\nu}} \in \cV$. For $1 \leq p \leq \infty$, we define the space $\ell^p(\Lambda;\cV)$ as the set of those sequences $\vec{v} = (v_{\bm{\nu}})_{\bm{\nu} \in \Lambda}$ for which $\|{\vec{v}}\|_{\cV,p} < \infty$, where 
\bes{
\|{\vec{v}}\|_{\cV,p} : = \left \{ \begin{array}{lc} \left ( \sum_{\bm{\nu} \in \Lambda} \|{v_{\bnu}}\|^p_{\cV} \right )^{1/p} & 1 \leq p < \infty 
\\ \sup_{\bm{\nu} \in \Lambda} \nm{v_{\bm{\nu}}}_{\cV} & p = \infty \end{array} \right . .
}
When $p = 2$ this is a Hilbert space with inner product
\[
\ip{\vec{u}}{\vec{v}}_{\cV,2} = \sum_{\bnu \in \Lambda}  \ip{u_{\bnu}}{v_{\bnu}}_{\cV} .
\]
Next, we define the weighted (Lebesgue-)Bochner space $L^p_{\varrho}(\cU;\cV)$
as the space consisting of (equivalence classes of) strongly $\varrho$-measurable functions $f: \cU \rightarrow \cV$ for which $\nm{f}_{L^{p}_{\varrho}(\cU ; \cV)} < \infty$, where 
\be{
\nm{f}_{L^p_{\varrho}(\cU;\cV)} : = 
\begin{cases} 
\left( \int_{\cU} \nm{f( \y)}_{\cV}^p \D \varrho (\y) \right)^{1/p} & 1 \leq p < \infty 
\\
\mathrm{ess} \sup_{\y \in \cU} \nm{f(\y)}_{\cV}  & p = \infty
\end{cases}
.
\label{L_p_U_V}
}
In general, we cannot work directly in the space $\cV$, since it is usually infinite dimensional. Hence, we consider a finite-dimensional discretization 
\begin{equation}
\label{eq:conforming}
\cV_h \subseteq \cV.
\end{equation}
Here $h > 0$ denotes a discretization parameter, e.g.\ the mesh size in the case of a finite element discretization. In the context of finite elements, assuming \eqref{eq:conforming} corresponds to considering so-called conforming discretizations. We let $\{ \varphi_k \}^{K}_{k=1}$ be a (not necessarily orthonormal) basis of $\cV_h$, where
$
K = K(h) = \dim(\cV_h).
$
We write $\cP_h: \cV \rightarrow \cV_h$
for the orthogonal projection onto $\cV_h$ and for $f \in L^2_{\varrho}(\cU ; \cV)$ we let $\cP_h f \in L^2_{\varrho}(\cU ; \cV_h)$ be the function defined almost everywhere as
\[
(\cP_h f)(\bm{y}) = \cP_h (f(\bm{y})), \quad \forall \y \in \cU.
\]

\subsection{DNNs}\label{Section_DNNs}

Let $f : \cU \rightarrow \cV$ and write its projection in terms of the basis $\{ \varphi_k \}^{K}_{k=1}$ for $\cV_h$ as
\bes{
f \approx (\cP_{h} f)(\bm{y}) = \sum^{K}_{k=1} c_{k}(\bm{y}) \varphi_k.
}
Notice that the \textit{coefficients} $c_k$ are scalar-valued functions of $\bm{y}$, i.e.\ $c_k : \cU \rightarrow \bbR$.
Our objective is to approximate the coefficients with a DNN. We consider standard feedforward DNN architectures of the form $\Phi : \bbR^d \rightarrow \bbR^K$, where
\be{
\label{Phi_NN_layers}
\Phi(\bm{y}) = \cA_{L+1} ( \sigma ( \cA_{L} ( \sigma ( \cdots \sigma ( \cA_0 (\bm{x}) ) \cdots ) ) ) ).
}
Here $\cA_l : \bbR^{N_{l}} \rightarrow \bbR^{N_{l+1}}$, $l = 0,\ldots,L+1$ are affine maps and $\sigma$ is the activation function, which we assume acts componentwise, i.e.\ $\sigma(\bm{y}) := (\sigma(y_i))^{d}_{i=1} $ for $\bm{y} = (y_i)^{d}_{i=1}$. The values $\{ N_l \}^{L+1}_{l=1}$ are the widths of the hidden layers. By definition $N_0 = d$ and $N_{L+2} = K$. Given \eqref{Phi_NN_layers}, we write
\be{
\label{fPhi_DNN}
f_{\Phi}(\bm{y}) = \sum^{K}_{k=1} (\Phi(\bm{y}))_k \varphi_k,
}
for the resulting approximation to $f$. We also write $\cN$ for a class of DNNs of the form \eqref{Phi_NN_layers}. We term $L$ the \textit{depth}, and denote this as $\mathrm{depth}(\cN)$. We write $\mathrm{param}(\cN)$ for the number of trainable parameters in $\cN$ (i.e.\ the number of weights and biases that parameterize $\cN$). We also write $\mathrm{size}(\cN)$ for the size of the DNNs in $\cN$. This is equal to the total number of nonzero weights and biases.

\subsection{Problem statement}

Let $f : \cU \rightarrow \cV$ be a Hilbert-valued function and consider $m$ sample points $\bm{y}_1,\ldots,\bm{y}_m$ drawn independently from the uniform measure \eqref{meas-unif}. We assume noisy evaluations of $f$ of the form
\be{
\label{samples_of_f}
d_i = f(\bm{y}_i) + n_i \in \cV_h,\quad \forall i = 1,\ldots,m,
}
where $n_i \in \cV$ is the $i$th noise term. It is important to note that the samples $d_i$ are elements of the finite-dimensional space $\cV_h$. Hence, the term $n_i$ encompasses the error involved in approximating $f(\bm{y}_i) \in \cV$ by an element of $\cV_h$. This takes into account the fact that these measurements are computed by some unspecified routine (e.g.\ a PDE solver in the case of parametric PDEs) which returns a value in the finite-dimensional Hilbert space $\cV_h$ (e.g.\ a finite element space). We do not specify precisely how this approximation is performed, nor how large this error is, but merely aim to show an error bound that scales linearly with respect to the $n_i$. A particular case is when $d_i = \cP_h(f(\bm{y}_i))$, but in what follows it is not necessary to assume this, as the numerical computation that leads to $d_i$ may not involve computing the projection $\cP_h$. We remark in passing that one can easily extend this analysis to the case where the space $\cV_h$ used for constructing the approximation $f_{\Phi}$ to $f$ differs from that used to compute the evaluation of $f$.

Since any algorithm for learning a DNN needs to take finite inputs, we assume the evaluations \eqref{samples_of_f} are provided to us via the basis $\{ \varphi_k\}$. To be precise, we assume we have access to the data
\be{
\label{data_we_get}
\left \{ d_{ik} \right \}^{m,k}_{i,k=1},\quad \mbox{where}\ f(\bm{y}_i) + n_i = \sum^{K}_{i=1} d_{ik} \varphi_k.
}
With this in hand, the problem is as follows:

\begin{problem}\label{prob:main}
Use the data \eqref{data_we_get} to learn a DNN $\hat{\Phi} : \bbR^d \rightarrow \bbR^K$ from the class $\cN$, and therefore an approximation $f_{\hat{\Phi}}$ to $f$ of the form \eqref{fPhi_DNN}.
\end{problem}

\section{Holomorphy, best \texorpdfstring{$s$} --term  polynomial approximation and hidden anisotropy}\label{s:holomorphyandstuff}

\subsection{Holomorphy}

We start by recalling the definition of holomorphy and of holomorphic extension for Hilbert-valued functions. The definition of holomorphy employed here is based on the notion of Gateaux partial derivative, although other equivalent definitions are possible (see \cite[Chapter 2]{herve2011analyticity}). The holomorphic extension assumption has played a crucial role in the context of parametric PDEs since the seminal work \cite{cohen2011analytic}; see also \cite{cohen2015high} and references therein.

\begin{definition}
[Holomorphy]
Let $\mathcal{O} \subseteq \mathbb{C}^{d}$ be an open set and  $\cV$ be a separable Hilbert space. A  function $f: \mathcal{O} \rightarrow \cV$ is holomorphic in $\mathcal{O}$ if and only if it is holomorphic with respect to each variable  in  $\mathcal{O}$. Equivalently, if and only if the following limit exists for any $z \in \mathcal{O}$ and any $j \in [d]$:
\[
\lim_{\substack{h \in \bbC \\ h \rightarrow 0}}  \dfrac{f(z+h \bm{e}_j)-f(z)}{h} \in \cV,\quad \mbox{where $\bm{e}_j = (\delta_{ij})^{d}_{i=1}$.}
\]
\end{definition}

\begin{definition}[Holomorphic extension]
\label{Assumption-holom}
Let $\cV$ be a Hilbert space. The function  $f: \cU \rightarrow \cV$ is \textit{holomorphic} in $\cU \subseteq \cO \subseteq \bbC^d$ if it has a holomorphic extension to $\cO$, i.e.\ there is a $\widetilde{f}: \mathcal{O} \rightarrow \cV$ that is holomorphic in $\mathcal{O}$ with  $\widetilde{f}|_{\cU}=f$. In this case, we also define $\nm{f}_{L^{\infty}(\cO; \cV)}:=\nm{{\widetilde{f}}}_{L^{\infty}(\cO; \cV)}$.
\end{definition}

We are interested in approximating Hilbert-valued functions $f : \cU \rightarrow \cV$ that admit a holomorphic extension to suitable open neighborhoods of $\cU$. Specifically, regions defined by \textit{Bernstein (poly)ellipses}.  When $d = 1$ the Bernstein ellipse of parameter $\rho > 1$ is defined by $\cE_\rho = \left\lbrace \tfrac{1}{2} (z +z^{-1}): z\in \bbC, 1 \leq |z|\leq \rho\right\rbrace \subset \bbC$. This is an ellipse with $\pm 1$ as its foci and major and minor semi-axis lengths given by $\frac12 (\rho\pm\rho^{-1})$. For $d \geq 1$, given $\brho = (\rho_j)^{d}_{j=1} \in \bbR^d$ with $\brho> \bm{1}$, we define the Bernstein polyellipse as
\[
\cE_{\brho} =\cE_{\rho_1} \times \cdots \times \cE_{\rho_d}  \subset \bbC^d.
\]
Further, we denote the class of unit-norm, Hilbert-valued functions that are holomorphic in $\cE_{\bm{\rho}}$ as
\bes{
\cB(\bm{\rho}) = \left \{ f : \cU \rightarrow \cV, \mbox{$f$ holomorphic in $\cE_{\bm{\rho}}$},\ \nm{f}_{L^{\infty}(\cE_{\bm{\rho}} ; \cV)} \leq 1\right \}.
}
Note that the parameter $\bm{\rho}$ dictates the smoothness of $f$ in the different coordinate directions. The larger $\rho_j$, the smoother $f$ is in the $j$th variable $y_j$. We say that $f$ has \textit{anisotropic} dependence on the variables $\bm{y}$ if the $\rho_j$ are potentially nonequal and that the anisotropy is \textit{hidden} if the $\rho_j$'s are unknown. 

\subsection{Polynomial approximation of holomorphic functions}\label{ss:polyapprox}

The approximation theory of functions that are holomorphic in Bernstein ellipses is a classical topic, especially in $d = 1$ dimensions. However, polynomial approximations have also proved to be extremely effective tools in $d \gg 1$ dimensions as well.

Let $\{ \Psi_{\nu} \}_{\nu \in \bbN_0}$ be the univariate orthonormal Legendre polynomial basis of $L^2_{\varrho}([-1,1])$ (see, for example, \cite{szego1975orthogonal}). Let $f \in \cB(\rho)$ for some $\rho >1$ and consider its expansion
\bes{
f = \sum_{\nu \in \bbN_0} c_{\nu} \Psi_{\nu},\qquad c_{\nu} :=\int_{-1}^1 f(y) \Psi_{\nu}(y) 2^{-1} \D y \in \cV,
}
where the coefficients $\{c_{\nu}\}_{\nu \in \bbN_0}$ are elements of $\cV$.
It is well-known that the truncated expansion $f_s = \sum^{s-1}_{\nu = 0} c_{\nu} \Psi_{\nu}$ converges to $f$ exponentially-fast with error $\cO(\sqrt{s}\rho^{-s})$ in $L^2_{\varrho}([-1,1];\cV)$.
\footnote{In the scalar-valued case, i.e.\ $\cV=\bbR$, this is due to classical estimates on the Legendre coefficients (see, e.g., \cite[Theorem 12.4.7]{davis1975interpolation}). A simple adaptation of this argument leads to the same result in the Hilbert-valued case.}

The situation becomes more complicated in $d \geq 2$ dimensions. Let $\{ \Psi_{\bm{\nu}} \}_{\bm{\nu} \in \bbN^d_0}$ be the tensor Legendre polynomial basis, defined by $\Psi_{\bm{\nu}} = \Psi_{\nu_1} \otimes \cdots \otimes \Psi_{\nu_d}$ for $\bm{\nu} = (\nu_k)^{d}_{k=1} \in \bbN^d_0$. Then, once more, we can then write $f \in \cB(\bm{\rho})$ as
\be{
\label{f_exp_Leg}
f = \sum_{\bm{\nu} \in \bbN^d_0} c_{\bm{\nu}} \Psi_{\bm{\nu}},\qquad c_{\bm{\nu}} := \int_{\cU} f(\bm{y}) \Psi_{\bm{\nu}}(\bm{y}) 2^{-d}\D \bm{y} \in \cV.
}
One aims to construct a polynomial approximation by selecting $s$ terms from this expansion, i.e.
\bes{
f \approx f_{S} =  \sum_{\bm{\nu} \in S} c_{\bm{\nu}} \Psi_{\bm{\nu}},
}
where $S \subset \bbN^d_0$ is a multi-index set with $|S| = s$. Unfortunately, standard, isotropic choices for $S$ such as the \textit{(isotropic) total degree} set $S = \{ \bm{\nu} = (\nu_k)^{d}_{k=1} : \nu_1 + \cdots + \nu_d \leq n \}$
generally lead to less favourable convergence rates in terms of $s$. In particular, when the dependence of $f$ on its variables $\bm{y}$ is anisotropic,  the index set $S$ may include many indices that correspond to small-norm coefficients $c_{\bm{\nu}}$.  This motivates an alternative approach based on \textit{best $s$-term approximation}, in which a polynomial approximation is formed by selecting the indices in \eqref{f_exp_Leg} corresponding to the largest $s$ of the coefficient norms $\nm{c_{\bm{\nu}}}_{\cV}$; see \cite{devore1998nonlinear,cohen2010convergence}.

Best $s$-term approximation is particularly well suited to approximating holomorphic functions with anisotropic dependence. If $f \in \cB(\bm{\rho})$ then it is known that, for any $\epsilon > 0$,
\be{
\label{best_s_term_poly}
\nm{f - f_s}_{L^2_{\varrho}(\cU ; \cV)} \leq \exp \left ( - \frac{1}{d+1}\left ( \frac{ s d! \prod^{d}_{j=1} \log(\rho_j) }{1+\epsilon} \right )^{1/d}  \right ),\qquad \forall s \geq \bar{s} = \bar{s}(d,\epsilon,\bm{\rho}),
}
where $f_s$ is the best $s$-term polynomial approximation to $f$ (see Theorem \ref{t:best_s_term_poly}). This shows that the best $s$-term approximation error converges exponentially-fast in $s^{1/d}$ without any prior knowledge of the $\rho_j$'s. In particular, as the right-hand side of \eqref{best_s_term_poly} depends on the product of the logarithms of the $\rho_j$'s, it is completely independent of their ordering. We note that, although a closed formula for $\bar{s}(d,\epsilon,\bm{\rho})$ in \eqref{best_s_term_poly} is not available, an inspection of the proof of Theorem~\ref{t:best_s_term_poly} (see Appendix~\ref{ss:polyapproxholobds}) and of \cite[Theorem~3.5]{opschoor2019exponential} reveals a constructive characterization of this quantity. 

It is notable that estimates such as \eqref{best_s_term_poly} say nothing about how to actually construct the approximation $f_s$ from sample values. Nevertheless, we consider the rate \eqref{best_s_term_poly} as the benchmark against which we compare the developed DNN procedure.

\begin{remark}
There are various results on the convergence rate of best $s$-term polynomial approximation of a holomorphic function.  Algebraic rates of convergence can be found in, for instance, \cite{cohen2011analytic,cohen2015high}. These are attractive since they also hold when $d = \infty$, thus theoretically permitting the approximation of functions of infinitely-many variables.  However, in finite dimensions the constants in these error bounds may be large \cite{tran2017analysis}. The bound \eqref{best_s_term_poly} is based on \cite[Sec.\ 3.9]{cohen2015high} and \cite{opschoor2019exponential}. However, the scaling $1/(d+1)$ is not sharp.  For \textit{quasi-optimal} error bounds, see \cite{beck2014convergence,beck2012optimal,tran2017analysis}.  The results of \cite{tran2017analysis} are asymptotically sharp as $s \rightarrow \infty$.  The challenge, however, for our purposes is that they do not generally lead to an approximation in a so-called lower set, unlike \eqref{best_s_term_poly}. This is an important component of our subsequent analysis (See \S \ref{ss:polyapproxholobds}).
\end{remark}

\subsection{Hidden anisotropy}

Motivated by this discussion, we now define the following class of functions:

\begin{definition}\label{def_class_f}
Let $d \geq 1$, $\gamma > 0$ and $\epsilon> 0$. Then $\mathcal{HA} = \mathcal{HA} (\gamma ,\epsilon,d)$ is the set of Hilbert-valued functions $f: \cU \rightarrow \cV$ that have a holomorphic extension to a Bernstein polyellipse $\cE_{\bm{\rho}}$ and satisfy $\nm{f}_{L^{\infty}(\cE_{\bm{\rho}} ; \cV)} \leq 1$, and where the parameters $\bm{\rho} = (\rho_j)^{d}_{j=1}$ satisfy
\begin{equation}\label{sigma_def}
\frac{1}{d+1}\left ( \frac{d! \prod^{d}_{j=1} \log(\rho_j) }{1+\epsilon} \right )^{1/d} \geq \gamma.
\end{equation}
\end{definition}
Observe that, thanks to \eqref{best_s_term_poly}, for this class of functions we have
\be{
\label{best_s_term_HA}
\nm{f - f_s}_{L^2_{\varrho}(\cU ; \cV)}
\leq \exp(-\gamma s^{1/d}),\quad \forall f \in \mathcal{HA}(\gamma,\epsilon,d),\; \forall s \geq \bar{s}(d,\epsilon,\bm{\rho}).
}
It is this rate we seek to obtain with the DNN approximation, with some polynomial scaling between $m$ and $s$. Specifically, we consider the following problem:

\begin{problem}\label{prob:main_holo}
Devise a DNN architecture and training procedure that solves Problem \ref{prob:main} and for which the error decays exponentially fast for all $f \in \mathcal{HA}(\gamma,\epsilon,d)$.
\end{problem}

\section{Main result}\label{s:mainres}

In this section we present our main result that resolves Problem 2. Its proof can be found in Appendix~\ref{ss:proof_main_thm}.
Given $m \geq 1$, $0 < \varepsilon < 1$ and $d \geq 1$, we first define $\cL = \cL(m,d,\varepsilon)$ as
\begin{equation}
\label{eq:def_log_factor}
\cL(m,d,\varepsilon) : = c_0 \cdot \log(2m) \cdot \left(  \log(2m) \cdot \min \{ \log(2m)+d ,  \log(2m) \cdot \log(2d) \}+ \log(\varepsilon^{-1} ) \right),
\end{equation}
where $c_0> 0$ is a universal constant, and 
\begin{equation}
\label{tildemdef}
\widetilde{m} = \widetilde{m}(m,d,\varepsilon) := m/\cL.
\end{equation}

\begin{theorem}\label{t:mainthm1}
There are universal constants $c_0,c_1,c_2,c_3 >0$ such that the following holds.
Let $d \geq 1$, $0 < \epsilon , \varepsilon < 1$, $\gamma > 0$, $m \geq 1$ and $\widetilde{m}$ be as in \eqref{tildemdef}. Let $\y_1, \ldots, \y_m$ be drawn independently from the uniform measure \eqref{meas-unif} on $\cU$. Then there is (a) a class of neural networks $\cN$ with ReLU activation functions, $N_0 = d$, $N_{L+2} = K$,
  \begin{align*}
 \begin{split}
 \mathrm{depth}(\cN)&  \leq c_1 \cdot  (1+ d\log(d)) \cdot  (1+ \log(\widetilde{m})) \cdot \left ( (\widetilde{m}/2^d)^{1/2}+ \log(\Delta) + \gamma \widetilde{m}^{1/(2d)} \right ),
  \\
  \mathrm{size}(\cN)&  \leq c_2 \cdot d\left( d (\widetilde{m}/2^d) +\left ((\widetilde{m} / 2^d)^{1/2}+ d \cdot \Delta \right ) \cdot\left( \log(\widetilde{m})+ \log(\Delta) + \gamma \widetilde{m}^{1/(2d)}\right)\right)+ K \cdot \Delta,  
 \end{split}
 \end{align*}
and $\mathrm{param}(\cN) \leq K \cdot \Delta$, where
  \[
 \Delta : =  \min \left \{ 2 \widetilde{m}^{3/2} 2^{d/2}, \E^2 (\widetilde{m}/2^d)^{1+\log(d)/(2\log(2))} , \frac{\widetilde{m}^{1/2}  \left ( \log(\widetilde{m}) + (d+1) \log(2) \right )^{d-1} }{2^{d/2-1} (d-1)!}\right \},
 \]
 (with the convention that $0! = 1$);
(b) a regularization function $\cJ : \cN \rightarrow [0,\infty)$ equivalent to a certain norm of the trainable parameters; and
(c) a choice of regularization parameter $\lambda$ involving only $\widetilde{m}$ and $d$; such that the following holds with probability at least $1-\varepsilon$. For all  $f \in \mathcal{HA}(\gamma,\epsilon,d)$
with noisy evaluations $d_i = f(\bm{y}_i) + n_i \in \cV_h$ as in \eqref{samples_of_f}, every minimizer $\hat{\Phi}$ of the training problem
\begin{equation}\label{trainingprob}
 \min_{\Phi \in \cN} \sqrt{\frac1m \sum^{m}_{i=1} \nm{f_{\Phi}(\bm{y}_i) - d_i }^2_{\cV} } +\lambda \cJ(\Phi),
\end{equation}
satisfies
\be{
\label{main_err_bd}
\| f- f_{\Phi}\|_{L^2_{\varrho}(\cU;\cV)} \leq c_3 \left( E_1 + E_2 + E_3 \right ),
}
for $\widetilde{m} \geq 2^{d} \bar{s}^2$, where $\bar{s} =\bar{s}(d,\epsilon,\bm{\rho})$ is as in \eqref{best_s_term_HA},  $f_{\Phi}$ is as in \eqref{fPhi_DNN} and, for  $\bm{e} = \frac{1}{\sqrt{m}} (n_i)^{m}_{i=1}$,
\be{
\label{E123_def}
E_1 = \exp (-\gamma\widetilde{m}^{1/(2d)} /\sqrt{2} ),
\quad
E_2 = \nm{\bm{e}}_{\cV,2},
\quad
E_3 = \nm{f - \cP_h(f)}_{L^{\infty}(\cU ; \cV)}.
}
\end{theorem}

This result asserts that there is a DNN architecture and training procedure with an explicit error bound comprising three terms. First, the \textit{approximation error} $E_1$. This quantifies how well $f$ is approximated by a DNN in terms of the number of samples $m$. It is exponentially small in $m^{1/(2d)}$, up to log terms. Furthermore, the log term $\cL = \cL(m,d,\varepsilon)$ effectively behaves roughly like $\log^3(m) \log(d)$, i.e.\ polylogarithmic in $m$ but only logarithmic in $d$. The second term is the \textit{measurement error} $E_2 = \nm{\bm{e}}_{\cV,2}$. This is, as discussed, the error in the pointwise evaluations of $f$ at the points $\bm{y}_i$. Note that we do not assume $\bm{e}$ stems from any random distribution -- the term $E_2$ implies that the noise can be adversarial, which is generally suitable in problems such as parametric PDEs, where the noise stems from the (deterministic) error incurred in the numerical PDE solve. The third term is the \textit{physical discretization error} $E_3$. Recall that we cannot work directly in the infinite-dimensional space $\cV$. This term accounts for the error induced by working in $\cV_h$ instead. Since the projection $\cP_h(f)(\bm{y})$ is the best approximation to $f(\bm{y}) \in \cV$ from $\cV_h$ in the $\cV$-norm, the term $E_3$ is optimal, up to a minor switch from the $L^2_{\varrho}(\cU ; \cV)$-norm on the left-hand side of \eqref{main_err_bd} to the $L^{\infty}(\cU ; \cV)$-norm on its right-hand side.

It is notable that the approximation error $E_1$ achieves a similar exponential rate of decay as the best $s$-term polynomial approximation error bound \eqref{best_s_term_HA}, except with $s$ replaced by $\sqrt{\widetilde{m}}$ and an additional factor of $1/\sqrt{2}$. Note that the former translates into $\widetilde{m} \propto s^2$, which is precisely the quadratic sample complexity that arises when computing $s$-sparse polynomial approximations from uniformly-distributed samples via compressed sensing \cite{chkifa2018polynomial,adcock2017compressed}. Hence, the presence of the term $E_1$ implies that the DNN procedure performs as well as current best-in-class polynomial approximation schemes based on compressed sensing.

As mentioned, the proof of Theorem \ref{t:mainthm1} is in fact based on re-interpreting such a polynomial-based compressed sensing procedure as a DNN training problem. This is done using a DNN to approximate the Legendre polynomial basis (using an approach due to \cite{opschoor2019exponential}), and therefore results in a class of DNN architectures $\cN$ in which only the weights in the final layer are trained, the remainder being fixed. As shown in step 2 of the proof of Theorem \ref{t:mainthm1}, the regularization $\cJ$ is a certain weighted $\ell^{2,1}$-norm over the weight matrix in the final layer. Note that the training problem may also seem unusual, since it involves fitting in the continuous $\nm{\cdot}_{\cV}$-norm. However, it can be reformulated as a discrete norm. Let $\bm{G} = \left ( \ip{\varphi_j}{\varphi_k}_{\cV} \right )^{K}_{j,k=1}$ be the Gram matrix of the basis $\{ \varphi_k \}^{K}_{k=1}$ of $\cV_h$.
Then \eqref{trainingprob} is equivalent to the problem $\min_{\Phi \in \cN} \sqrt{\frac1m \sum^{m}_{i=1} \nm{\Phi(\bm{y}_i) - \bm{d}_i }^2_{\bm{G},2}} + \lambda\cJ(\Phi)$,
where $\bm{d}_i = (d_{ik})^{K}_{k=1} \in \bbR^K$ and $\nm{\bm{c}}_{\bm{G},2} = \sqrt{\ip{\bm{G} \bm{c}}{\bm{c}}}$, which involves a standard (weighted) $\ell^2$-norm  over $\bbR^K$.
However, we also note that the loss function in \eqref{trainingprob} is is rather different from the standard $\ell^2$-loss used commonly in DL, since the data fidelity term is not squared. The reason for this, as discussed in \S \ref{ss:HilbertCS}, is to ensure property (c). If the loss term were squared, the optimal choice regularization parameter $\lambda$ would depend on the unknown error terms $E_1$, $E_2$ and $E_3$. It is rather remarkable that forgoing the squaring leads to such a desirable property for $\lambda$. We are not aware of any existing results in DNN training where the regularization parameter can be set optimally in a function-independent way.

\section{Numerical exploration}

As commented previously, since Theorem \ref{t:mainthm1} emulates an existing polynomial-based method with a DNN, the resulting training procedure is not expected to give better results in practice.
We now present preliminary experiments on the efficiency of training DNNs for approximating solutions to parametric PDEs from limited data. A more comprehensive study of the efficacy of DL techniques for problems in computational UQ will be presented in an upcoming work. Unlike in Theorem \ref{t:mainthm1}, we consider fully-connected DNNs trained using standard (unregularized) loss functions. In addition, differing from previous works -- including  \cite{geist2020numerical}, which focused on the scaling of accuracy with the dimension $d$ in the presence of large amounts of training data -- in this study we investigate the trade-off between accuracy and the number of samples $m$.
We also include a comparison to the best-in-class SCS scheme \cite{dexter2019mixed} (see \S \ref{ss:previous_work} and \S \ref{app:SCS}). 

\subsection{Setup}

We approximate a parametric elliptic PDE defined over the spatial domain $\Omega= (0,1)^2$ and parametric domain $\mathcal{U}= [-1,1]^d$ with the uniform probability measure \eqref{meas-unif}. Specifically, given $g \in L^2(\Omega)$ we seek a function $u: \Omega \times \mathcal{U} \rightarrow \mathbb{R} $ satisfying
\be{
\label{PDE}
-\nabla \cdot (a(\x,\y) \nabla u(\x,\y)) =g(\x),\  \forall \x \in \Omega,\, \y \in \cU,\quad u(\x,\y)= 0,\ \forall \x \in \partial \Omega,\, \y \in \cU.
}
We choose the function $g$ to be independent of $\y$ for simplicity.
We use a finite element method for spatial discretization, based on a triangulation with $K = 1089$ nodes and meshsize $h = \sqrt{2}/32$. In this study we consider only the error in approximating the parametric PDE, therefore we use the same finite element discretization in generating all of the sample training and testing data, and also in discretization and solution of the SCS problem. See \S \ref{app:FEM} for further details.

The training data for the DNNs and for the SCS method is generated by solving \eqref{PDE} at a set of uniform random points $\{\y_i\}_{i=1}^{m_{\max}} \subset \cU$, yielding a set of solutions $\{u_h(\cdot,\y_i)\}_{i=1}^{m_{\max}}$, where $u_h(\cdot,\bm{y}) \in \cV_h$ is the computed solution of \eqref{PDE}. Here we examine both the efficiency of DL techniques on fixed subsets of training data of size $m \le m_{\max}$ and the effect of increasing $m$ up to $m_{\max}$ on the accuracy of the approximations.
We also examine the average performance of both methods, running multiple trials training the DNNs and solving the SCS problem given different sets of uniform random points generated with the trial number as the seed.
The testing data is generated in the same way by evaluating \eqref{PDE} at a set of points $\{\y_i\}_{i=1}^{m_{\textnormal{test}}}$. However instead of using random points to test, we use a deterministic high-order sparse grid stochastic collocation method to generate the set of testing points and data $\{u_h(\cdot,\bm{y}_i)\}_{i=1}^{m_{\textnormal{test}}}$ (see \cite{Nobile2008}).
The sparse grid method is selected due to the superior convergence over standard Monte Carlo integration in evaluating the global testing error metrics, chosen here to be the Bochner norms $L^2_\varrho(\cU;L^2(\Omega))$ and $L^2_\varrho(\cU;H_0^1(\Omega))$ (see \eqref{L_p_U_V}). See \S \ref{app:testerr} for further details. 

The DNN strategy is based on \S \ref{Section_DNNs}. For $\y \in \cU$, we define the Finite Element (FE) and DNN approximations by
\be{
\label{FE_DNN_approx}
 u_h(\bm{y}) =\sum^{K}_{k=1} c_{k}(\bm{y}) \varphi_k \qan u_{\Phi,h}(\y) = \sum^{K}_{k=1} (\Phi(\bm{y}))_k \varphi_k,\ \Phi : \bbR^d \rightarrow \bbR^K,
}
respectively. 
We consider several types of fully-connected DNN architectures with input dimension $N_0=d$, output dimension $N_{L+2}=K$, and constant hidden-layer widths, i.e.\ $N_1=N_2=\ldots=N_{L+1}=M$ for some $M\in\bbN$. We follow the convention from \cite{adcock2020gap} of referring to a DNN with activation function $\sigma$, $L$ hidden layers and $M$ nodes per hidden layer as a $\sigma$ $L\times M$ DNN. 
In this study we focus on either the $\tanh$, $\rho(x) = \tanh(x)$, the ReLU, $\rho(x) = \max\{x, 0\}$, or the Leaky-ReLU, $\rho(x) = \max\{x, 0.2 x\}$, activation functions.\footnote{As noted in \cite{geist2020numerical}, the Leaky-ReLU can help avoid the occurrence of `dead neurons' in training. Further, \cite[Remark 3.3]{geist2020numerical} implies that theoretical guarantees such as Theorem \ref{t:mainthm1} also hold if the ReLU is replaced by the Leaky-ReLU, up to a minor change in the architecture of the family of DNNs $\cN$.}
For details on  the training approach, see \S \ref{app:training}.

\subsection{Effective architectures and loss functions and efficiency of training}

\begin{figure}[t]
\centering
\includegraphics[width=0.30\paperwidth, clip=true, trim=0mm 0mm 0mm 0mm]{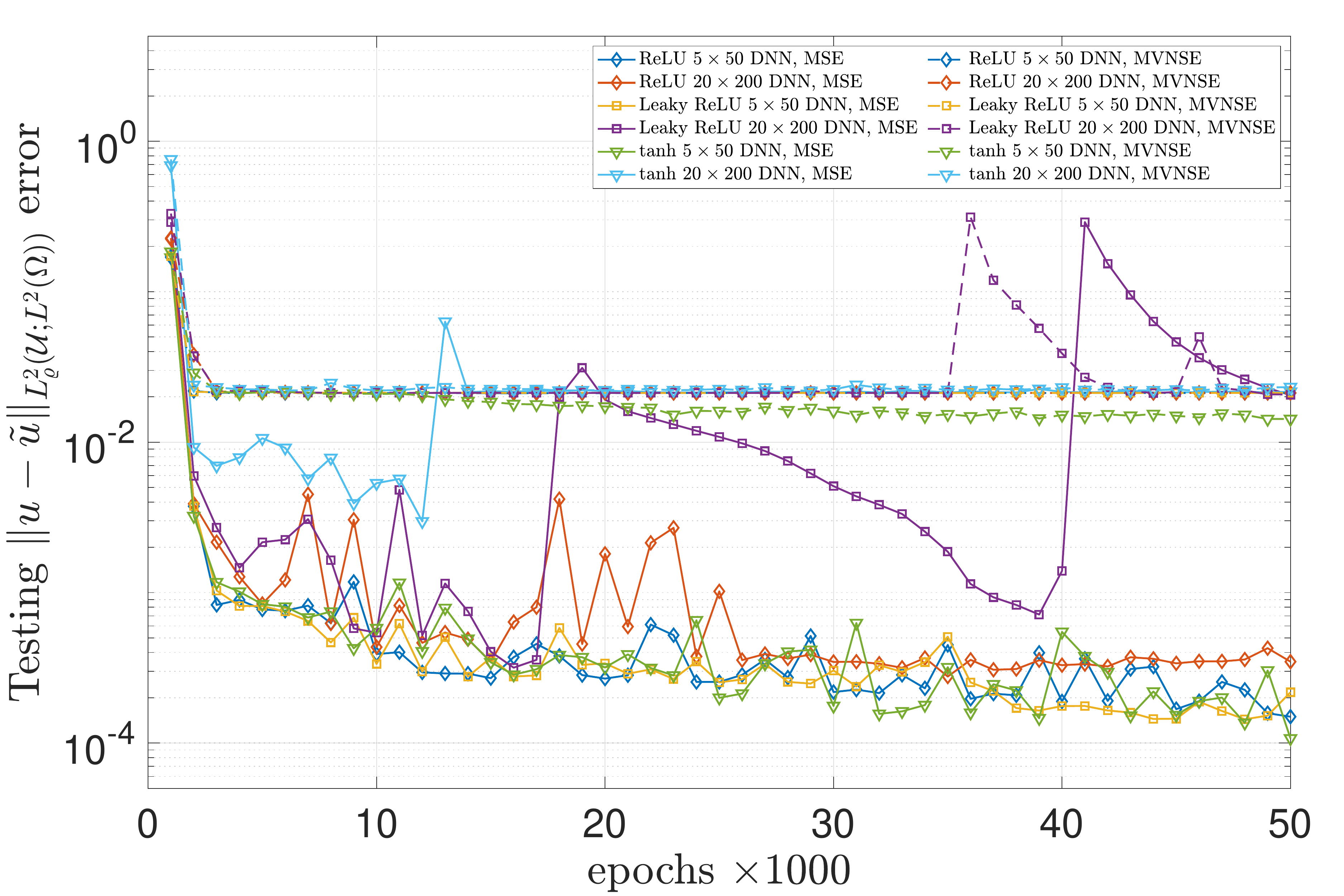}\qquad
\includegraphics[width=0.30\paperwidth, clip=true, trim=0mm 0mm 0mm 0mm]{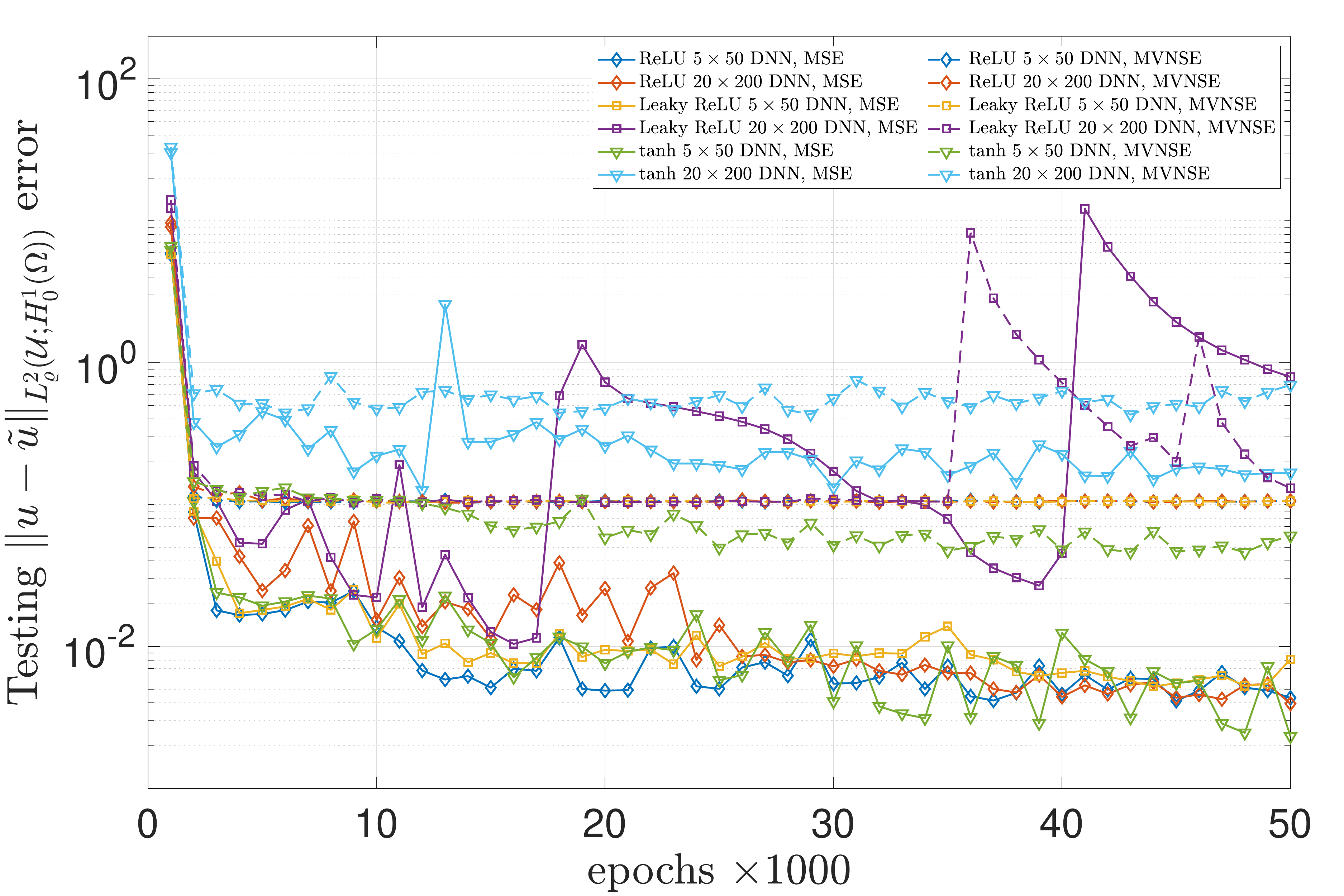}\\
\includegraphics[width=0.30\paperwidth, clip=true, trim=0mm 0mm 0mm 0mm]{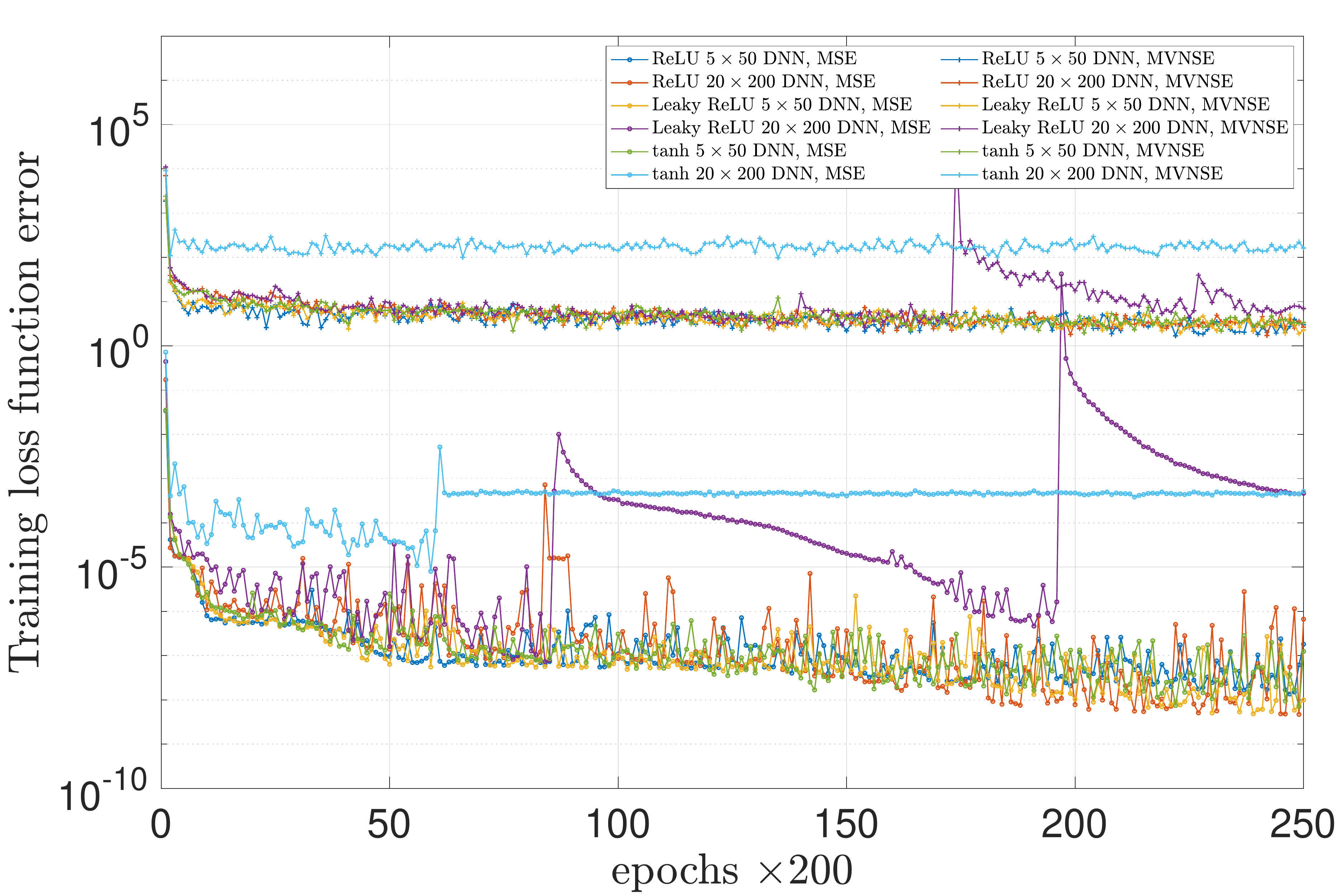}\qquad
\includegraphics[width=0.30\paperwidth, clip=true, trim=0mm 0mm 0mm 0mm]{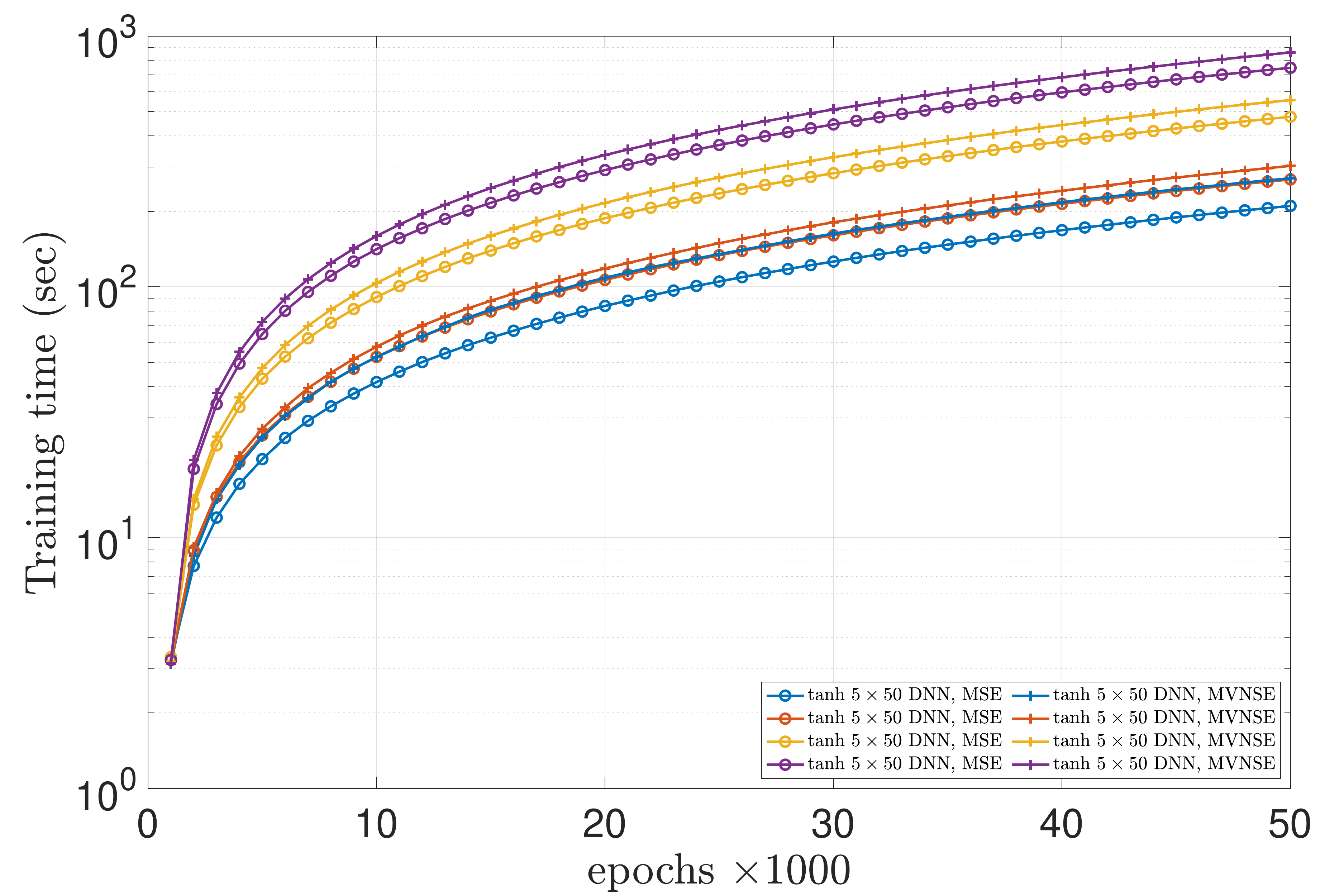}

\caption{Comparison of \textbf{(top-left)} testing error in $L^2_\varrho(\cU; L^2(\Omega))$, \textbf{(top-right)} testing error in $L^2_\varrho(\cU; H_0^1(\Omega))$, and \textbf{(bottom-left)} training error for a variety of DNNs trained with MSE loss function \eqref{MSE_loss} and MVNSE loss \eqref{H1_loss} on $m=400$ samples, and \textbf{(bottom-right)} training time of a $\tanh$ $5\times 50$ DNN with both loss functions and a range of samples.
}
\label{Example1fig}
\end{figure}

\begin{figure}[t]
\centering
\includegraphics[width=0.232\paperwidth,clip=true,trim=0mm 0mm 0mm 0mm]{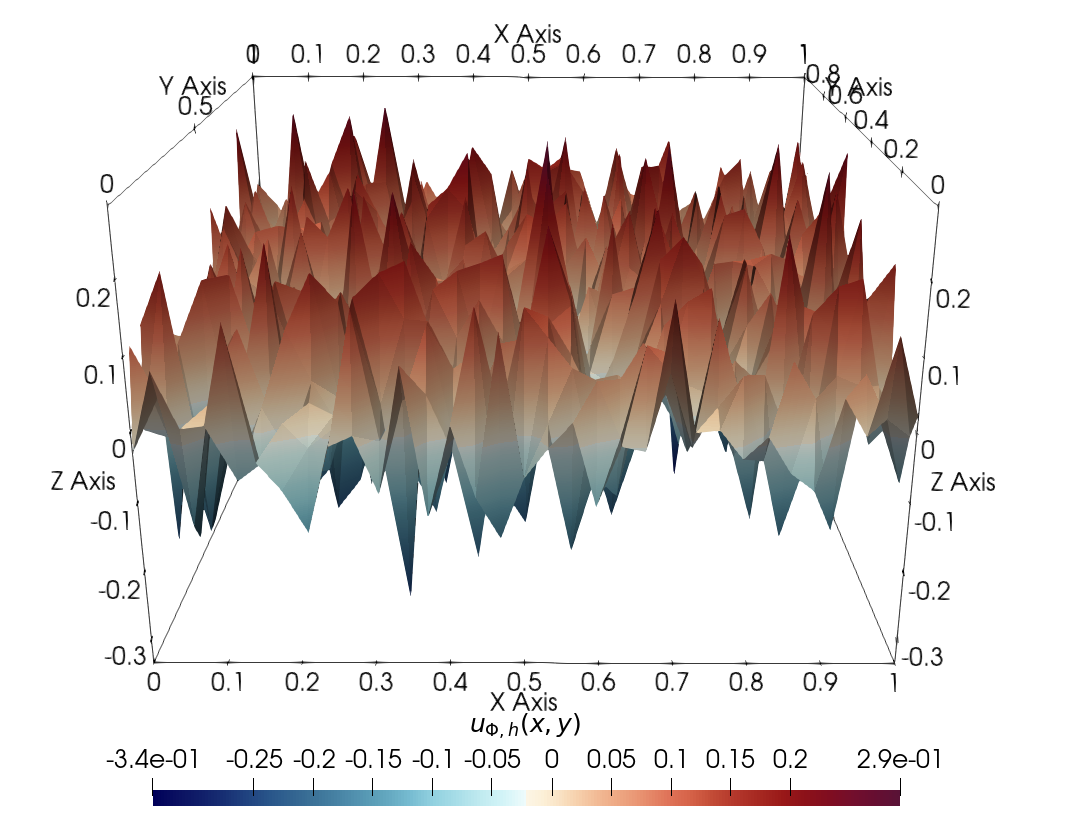}
\includegraphics[width=0.232\paperwidth,clip=true,trim=0mm 0mm 0mm 0mm]{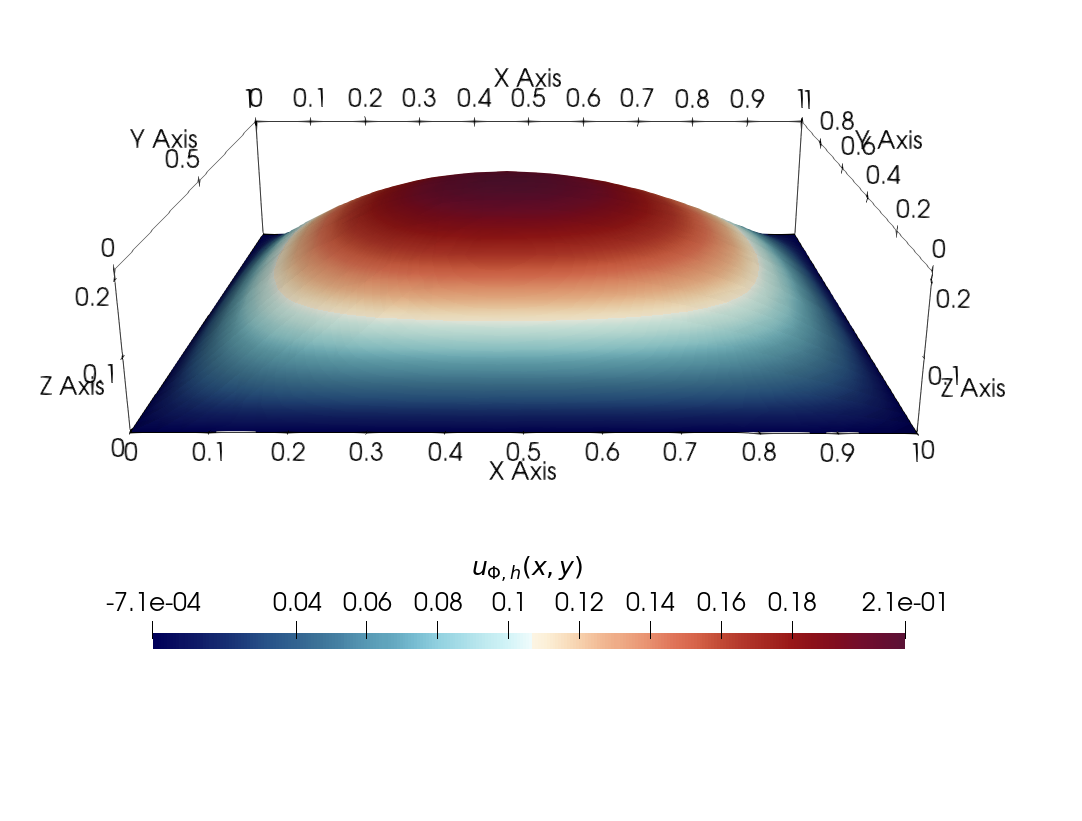}
\includegraphics[width=0.232\paperwidth,clip=true,trim=0mm 0mm 0mm 0mm]{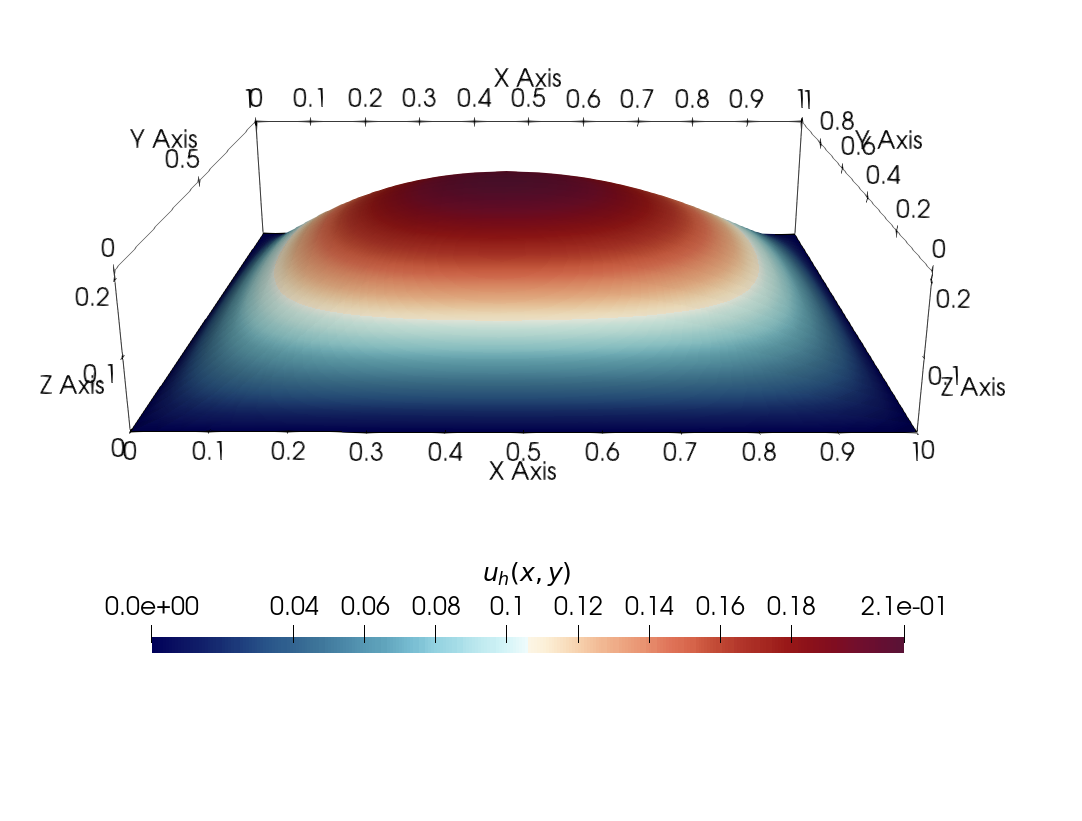}
\caption{Prediction for $u_h(\x,\y)$ from a $\mathrm{tanh}$ $5 \times 50$ DNN at $\y=[0.995184,0]^{\top}$ \textbf{(left)} after $2$ epochs of Adam (MSE $6.4255$) and \textbf{(middle)} after training for $2045$ epochs (MSE $4.879 \cdot 10^{-7}$), and \textbf{(right)} the reference FE solution. At this $\y$, $\|u_h(\y) - u_{\Phi,h}(\y)\|_{L^2(\Omega)} = 8.417 \cdot 10^{-4}$ and $\|u_h(\y) - u_{\Phi,h}(\y)\|_{H^1(\Omega)} = 2.315\cdot 10^{-2}$. 
}
\label{Example2fig}
\end{figure}

We first study the performance of the DL approach based on two different loss functions. The first loss function is the well-known mean square error loss, given in terms of the coefficients \eqref{FE_DNN_approx} as
\begin{equation}
\label{MSE_loss}
\textnormal{MSE}(\bm{y}) := {\dfrac{1}{m}\sum_{i=1}^m} \sum^{K}_{k=1} {(c_{k}(\bm{y}_i)  - (\Phi(\bm{y}_i))_k )^2}.
\end{equation}
On the other hand, motivated by the recent work \citep{geist2020numerical} we also consider the squared $\cV$-norm loss function, which incorporates information about the complexity of the domain in the training procedure. This is defined by 
\begin{equation}
\label{H1_loss}
\textnormal{MVNSE}(\y) := \frac{1}{m} \sum_{i=1}^{m} \| u_h(\y_i) - u_{\Phi, h}(\y_i) \|_\cV^2.
\end{equation}
In Figure \ref{Example1fig}, we study the efficacy of minimizing either \eqref{MSE_loss} or \eqref{H1_loss} given solution data from problem \eqref{PDE} with a simple affine coefficient given by
\begin{equation}
\label{affine_coef}
a(\bm{x}, \bm{y})= 3 + x_1 y_1 + x_2 y_2.
\end{equation}
We make several observations. 
First, when comparing the training time, identical networks trained with loss \eqref{H1_loss} take longer to complete 50,000 epochs of training than those trained with loss \eqref{MSE_loss}, and the overall time scales poorly with increasing training set size $m$ and increasing number of FE basis elements $K$ for both loss functions.
Second, DNN architectures trained with loss function \eqref{H1_loss} underperform identical architectures trained with the MSE loss in both testing metrics.
However, it is interesting that the difference is largest in the case of the $L^2_\varrho(\cU;L^2(\Omega))$ testing error, where the best-performing $\tanh$, ReLU, and Leaky-ReLU $5\times 50$ DNNs trained with the MSE loss achieved order $10^{-3}$ error. 
Third, we note the non-monotonic decrease in training error for both loss functions, and stagnation in error of some of the larger networks.
Figure \ref{Example2fig} displays a visualization comparing solutions output by the DNN at a fixed $\y$ both early in the training and with a much improved result after achieving order $10^{-6}$ error in the MSE loss.

\subsection{Sample complexity of training}

In this section we run a large-scale study of the average testing errors and training times over a range of 10 trials, with increasing number of training samples $m$ up to $m_{\max} = 675$, and testing with a sparse grid quadrature rule with $m_{\textnormal{test}} = 1861$ testing points. 
For these experiments, we consider a modification of the example from \cite{Nobile2008} of a diffusion coefficient with one-dimensional (layered) spatial dependence given by
\begin{align*}
a(\bm{x},\y) 
& = \exp\left(1+ y_1 \left(\frac{\sqrt{\pi}\beta}{2}\right)^{1/2} + \sum_{i=2}^d \; \zeta_i \; \vartheta_i(\bm{x}) \; y_i\right) \numberthis \label{log_KL_example} \\
\zeta_i 
& := (\sqrt{\pi} \beta)^{1/2} \exp\left( \frac{-\left( \left\lfloor \frac{i}{2} \right\rfloor \pi \beta\right)^2}{8} \right), 
\quad
\vartheta_i(\bm{x}) 
 := \left\{ \begin{array}{rl} \sin\left( \left\lfloor\frac{i}{2}\right\rfloor \pi x_1/\beta_p \right),  \text{ if $i$ is even,} \\[0.1cm] \cos\left(\left\lfloor\frac{i}{2}\right\rfloor \pi x_1/\beta_p \right),  \text{ if $i$ is odd.} \end{array} \right. .
\end{align*}
Here we let $\beta_c = 1/8$, and $\beta_p = \max\{1, 2 \beta_c\}$, $\beta = \beta_c/\beta_p$. We consider this problem with parameter dimension $d=30$.

 \begin{figure}[t]
\centering
\includegraphics[width=0.22\paperwidth,clip=true,trim=0mm 0mm 0mm 0mm]{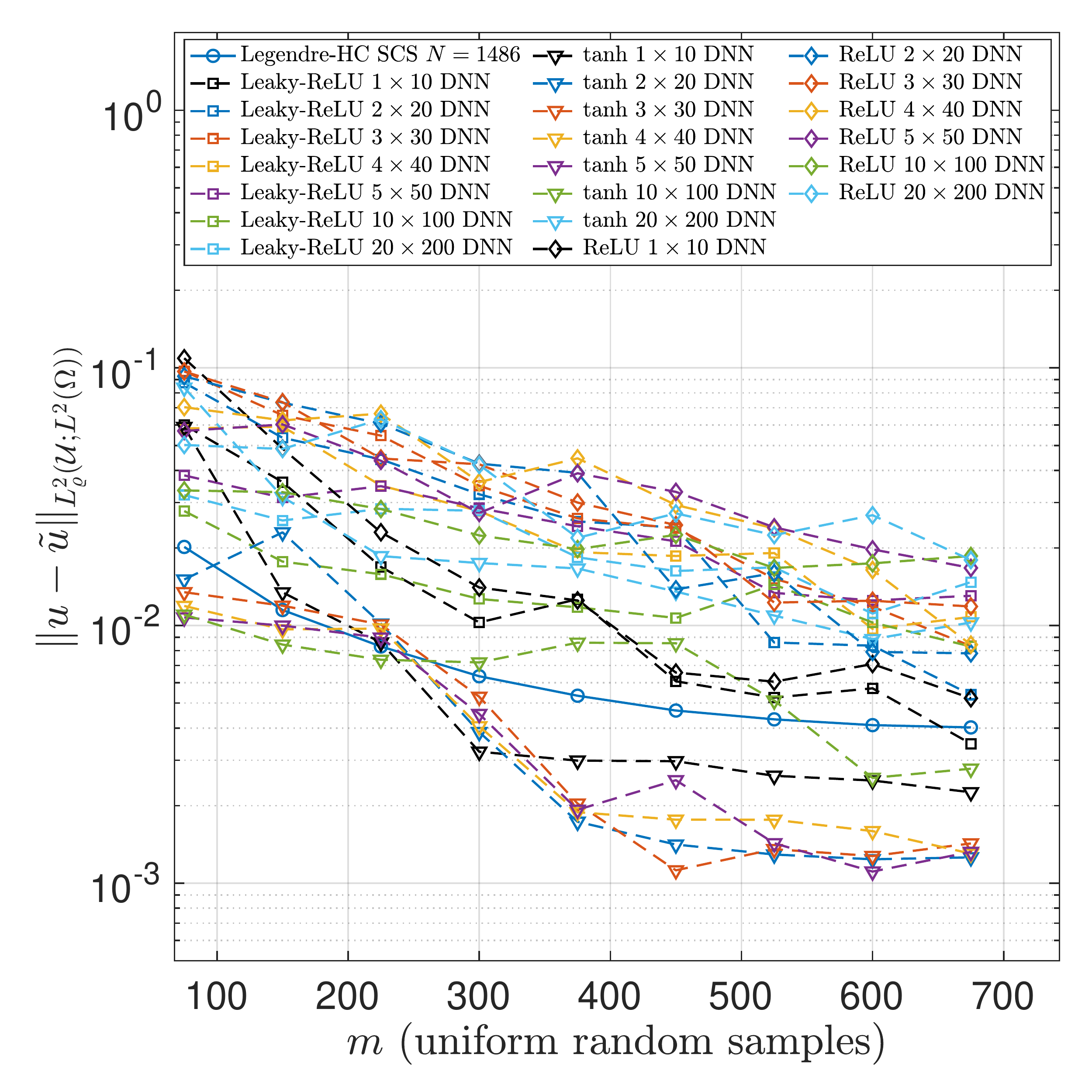}
\includegraphics[width=0.22\paperwidth,clip=true,trim=0mm 0mm 0mm 0mm]{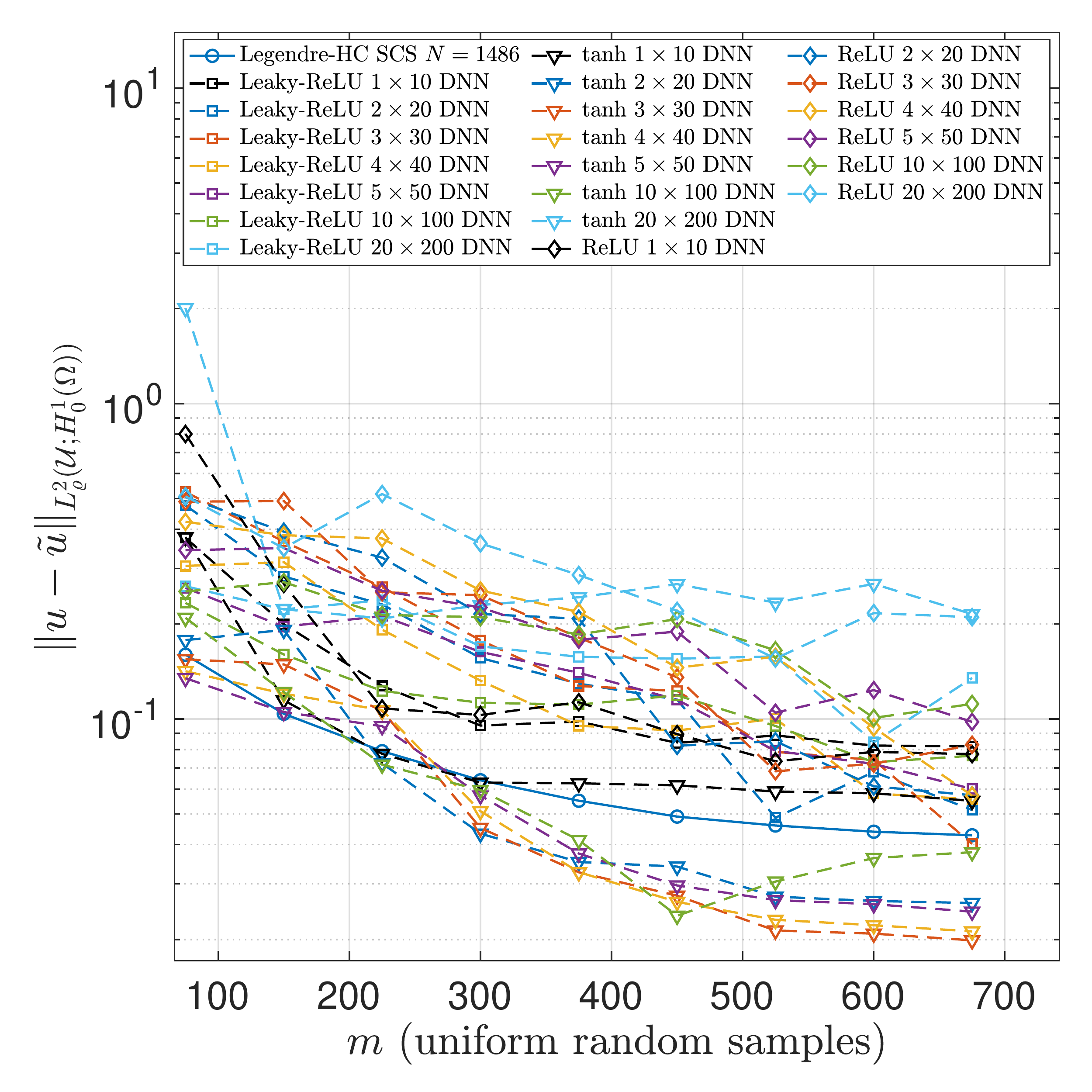}
\includegraphics[width=0.22\paperwidth,clip=true,trim=0mm 0mm 0mm 0mm]{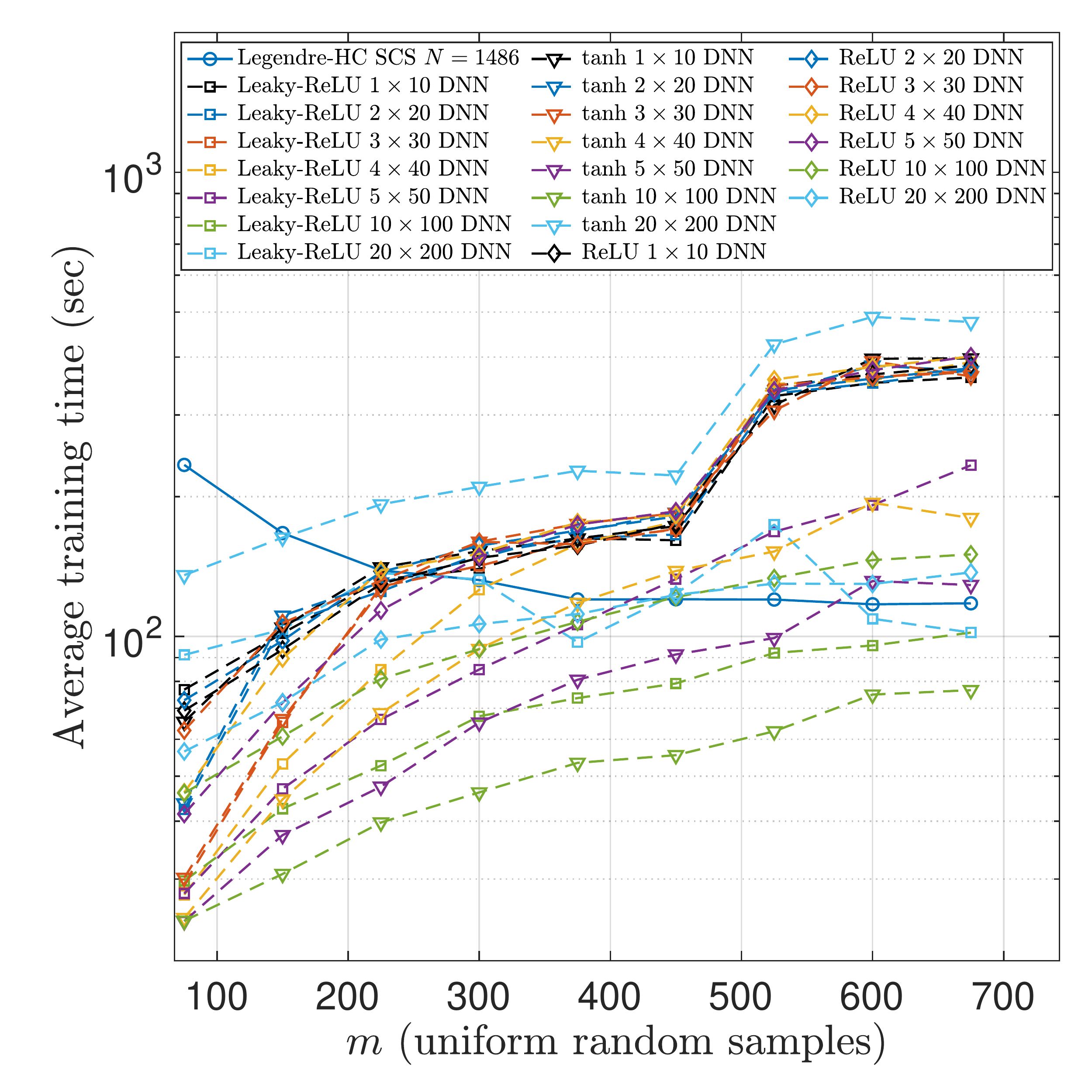}
\caption {Comparison of average testing errors in {\bf(left)} $L^2_\varrho(\cU; L^2(\Omega))$ and {\bf(middle)} $L^2_\varrho(\cU; H_0^1(\Omega))$, and {\bf(right)} average training times of the SCS method and various DNN architectures in solving problem \eqref{PDE} with coefficient \eqref{log_KL_example} in $d=30$ dimension.}
\label{Example3fig}
\end{figure}

Figure \ref{Example3fig} displays the result solving problem \eqref{PDE} with coefficient \eqref{log_KL_example} with SCS and a variety of DNN architectures. For the SCS method, we use the Legendre basis and hyperbolic cross of order $p=6$ with cardinality $N=1486$. The DNN architectures have depth parameter $L$ and number of hidden layer nodes $M$ chosen so that the ratio $L/M =0.1$. 
Due to the non-monotonic decrease in error during training, in this work we employ checkpointing to ensure the parameters achieving the lowest loss are saved and later reloaded for testing, as described in Section \ref{app:training}.
In testing we observe competitive performance using DNNs with the ReLU and Leaky-ReLU activation function. 
We also observe superior performance over SCS and ReLU and Leaky-ReLU DNNs using DNNs with $\tanh$ activation function, with such networks achieving testing errors on average approximately 3.2 times lower in the $L^2_\varrho(\cU;L^2(\Omega))$ error and 2.15 times lower in the $L^2_\varrho(\cU;H_0^1(\Omega))$ error than the SCS method given 675 training samples.
We also include a comparison of the average training time of the SCS and DNN approaches. While the average times are overall quite similar between the SCS method and training the DNNs, we note that the 
DNNs are trained on a GPU with accelerated matrix-vector product operations. Therefore this comparison does not provide a good estimate of computational complexity. We leave a study of the computational efficiency of DL techniques for such problems to a future work.

\section{Conclusion}

In this paper, we first established a novel theoretical result, Theorem \ref{t:mainthm1}, asserting the existence of a DNN architecture and training procedure that performs as well as current best-in-class schemes. While this theorem does not explain the success of standard architectures and training, it does highlight the potential of DNNs for holomorphic function approximation. These preliminary numerical results shown above also indicate this promise. The architecture and training differ from that described in Theorem \ref{t:mainthm1}, since the networks are much shallower and the loss function simpler. Yet, through a suitable choice of architecture and optimizer, we are not only able to match such performance of current best-in-class schemes using a simpler setup, but also outperform it in this example. These preliminary results indicate practical promise of the DNN approach, as well as the need for further improvements to the theory so as to address more realistic training scenarios. Results of a more comprehensive study will be reported in a future work.

\bibliography{msml21-ADBM-refs.bib}

\appendix

\section{Further details on the experiments}

\subsection{Finite element discretization}\label{app:FEM}

For the spatial discretization we rely on the finite element method as implemented by Dolfin \citep{LoggWells2010a}, and accessed through the python FEniCS project \citep{AlnaesBlechta2015a}. We generate a regular triangulation $\cT_h$ of $\overline{\Omega}$ composed of triangles $T$ of equal diameter $h_T = h$. We consider a conforming discretization, meaning a finite dimensional subspace $\cV_h \subset \cV := H_0^1(\Omega)$, where $\cV_h$ is chosen as the usual Lagrange Finite Elements (FE) of order $k=1$. 
We rely on the Dolfin {\tt UnitSquareMesh} method to generate a mesh with 33 nodes per side, corresponding to a finite element triangulation with $K=1089$ nodes, 2048 elements and meshsize $h = \sqrt{2}/32$.
 See \citep{dexter2019mixed} for further implementation details.

\subsection{Testing error}\label{app:testerr}

As discussed, rather than using random points to evaluate the test error, we use a high-order stochastic collocation reference solution (see \cite{Nobile2008}) based on a deterministically-generated quadrature rule. Specifically, we apply a Smolyak sparse-grid quadrature rule based on Clenshaw-Curtis points to generate the testing data with the TASMANIAN software package \cite{doecode_6305,stoyanov2015tasmanian}. In testing, we choose the level $\ell$ of the sparse grid rule such that $m_{\textnormal{test}} \gg m_{\max}$. 
The testing error is recorded in the Bochner norms $L^2_\varrho(\cU;L^2(\Omega))$ and $L^2_\varrho(\cU;H_0^1(\Omega))$ (see \eqref{L_p_U_V}) and approximated as the square root of the result of the quadrature formulas
\begin{align*}
\sum_{i=1}^{m_{\textnormal{test}}} \| u_h(\y_i) - \tilde{u}_h(\y_i) \|_{L^2(\Omega)}^2 w_i 
\qan 
\sum_{i=1}^{m_{\textnormal{test}}} \| u_h(\y_i) - \tilde{u}_h(\y_i) \|_{H_0^1(\Omega)}^2 w_i,
\end{align*}
respectively, where $w_i$ are the quadrature weights associated with the sparse grid rule and $\tilde{u}_h$ are the approximations obtained with either the DNNs or SCS. When testing multiple trials, we report the average of the errors over all of the trials.

\subsection{DNN training}\label{app:training}

We follow a similar training methodology to that of \cite{adcock2020gap}. All weights and biases of the DNN are initialized as normal random variables with mean 0 and variance 0.01, with the same seed 0 for each network. We perform calculations in single precision, using the Adam optimizer \cite{kingma2014adam} with an exponentially-decaying learning rate and training for 50,000 epochs or until a stopping tolerance of $5\times 10^{-7}$ is met. 
We also employ checkpointing, saving the DNN weights and biases once the error has decreased to $1/16$th the error of the previous checkpoint, and keeping the configuration which provided the lowest training error.
Due to the large size of the training data, we use a batch size of $\min\{m, 256\}$ for each training set of size $m$. 
As discussed, the loss function is chosen either as the MSE loss \eqref{MSE_loss} or the MVNSE loss \eqref{H1_loss}.

\subsection{The SCS method}\label{app:SCS}

The SCS method uses multivariate orthonormal Legendre polynomial approximation as described in \S \ref{ss:polyapprox} and Appendix \ref{ss:vector_recovery}, with the hyperbolic cross index set $\Lambda = \Lambda_{s-1}^{\textnormal{HC}}$ as in \eqref{HC_index_set}. Given the measurement matrix $\bm{A}\in \mathbb{R}^{m\times N}$ from \eqref{def-measMatrix} and measurement vector $\bm{b} = \frac{1}{\sqrt{m}} (u_h(\y_i))_{i=1}^m \in \cV_h^m$, we solve the LASSO problem
\be{
\label{LASSOproblem}
\min_{\bm{z} \in \cV^N_h} \lambda \nm{\bm{z}}_{\cV,1} + \nm{\bm{A} \bm{z} - \bm{b} }^2_{\cV,2},
}
using a combination of Bregman iterations \cite{YOGD08} and fixed point continuation for ISTA \cite{HYZ08}. The full implementation details, including choice of parameters for the solvers and value of $\lambda$, can be found in \cite{dexter2019mixed}.

\section{Proof of Theorem \ref{t:mainthm1}}

\subsection{Formulation as a vector recovery problem}\label{ss:vector_recovery}

Following \S \ref{ss:polyapprox}, we first consider approximating $f$ using the tensor Legendre polynomial basis $\{ \Psi_{\bm{\nu}} \}_{\bm{\nu} \in \bbN^d_0}$. Write
\bes{
f = \sum_{\bm{\nu} \in \bbN^d_0} c_{\bm{\nu}} \Psi_{\bm{\nu}},\qquad c_{\bm{\nu}} =  \int_{\cU} f(\bm{y}) \Psi_{\bm{\nu}}(\bm{y}) 2^{-d}\D \bm{y} \in \cV,
}
and let $\bm{c} = (c_{\bm{\nu}})_{\bm{\nu} \in \bbN^d_0} \in \ell^2(\bbN^d_0 ; \cV)$ be the infinite vector of coefficients of $f$.
Fix $s \in \bbN$ and let
\be{
\label{HC_index_set}
\Lambda = \Lambda^{\mathrm{HC}}_{s-1} = \left \{ \bm{\nu} = (\nu_k)^{d}_{k=1} \in \bbN^d_0 : \prod^{d}_{k=1} (\nu_k+1) \leq s \right \} \subset \bbN^d_0,
}
be the hyperbolic cross index set of index $s-1$. Let $N = |\Lambda|$ and $\bm{\nu}_1,\ldots,\bm{\nu}_N$ be an indexing of the multi-indices in $\Lambda$ and define  the normalized measurement matrix.
\begin{equation}\label{def-measMatrix}
\bm{A} =\left(\frac{\Psi_{\bnu_j} (\y_i)}{\sqrt{m}} \right)^{m,N}_{i,j=1} \in \bbR^{m \times N}.
\end{equation}
We also define the the normalized measurement and error vectors
\[
\bm{b} = \frac{1}{\sqrt{m}} \left ( f(\bm{y}_i) + n_i \right )^{m}_{i=1} \in \cV^m_h,\qan \bm{e} = \frac{1}{\sqrt{m}} (n_i)^{m}_{i=1} \in \cV^m.
\]
Now define
\be{
\label{f_exp_trunc}
f_{\Lambda} = \sum_{\bm{\nu} \in \Lambda} c_{\bm{\nu}} \Psi_{\bm{\nu}},
}
as the truncated expansion of $f$ based on the index set $\Lambda$ and
\bes{
\bm{c}_{\Lambda} = (c_{\bm{\nu}_j})^{N}_{j=1} \in \cV^N,
}
as the finite vector of coefficients of $f$ with indices in $\Lambda$.
Then we have
\[
\bm{A} \bm{c}_{\Lambda} = \frac{1}{\sqrt{m}} \left ( f_{\Lambda}(\bm{y}_i) \right )^{m}_{i=1} =  \frac{1}{\sqrt{m}} (f(\bm{y}_i))^{m}_{i=1} - \frac{1}{\sqrt{m}} \left ( f(\bm{y}_i) - f_{\Lambda}(\bm{y}_i) \right )^{m}_{i=1},
\]
and therefore
\begin{equation}\label{linsys_for_cLambda}
\bm{A} \bm{c}_{\Lambda} + \bm{e} + \bm{e'} = \bm{b},
\end{equation}
where
\[
\bm{e'} = \frac{1}{\sqrt{m}}  \left ( f(\bm{y}_i) - f_{\Lambda}(\bm{y}_i) \right )^{m}_{i=1}.
\]
Hence, the recovery of the coefficients $\bm{c}_{\Lambda}$ of $f$ is equivalent to solving the noisy linear system of equations \eqref{linsys_for_cLambda}.

\subsection{Hilbert-valued compressed sensing}\label{ss:HilbertCS}

To solve this, we consider techniques from \textit{Hilbert-valued} compressed sensing \cite{dexter2019mixed}. In classical (scalar-valued) compressed sensing, one aims to solve \eqref{linsys_for_cLambda} by finding a solution with minimal $\ell^1$-norm. In Hilbert-valued compressed sensing, we use the $\ell^1(\Lambda ; \cV)$-norm instead. Both classical and Hilbert-valued compressed sensing are usually formulated as either a quadratically-constrained basis pursuit or LASSO problem. However, we instead consider the \textit{Square Root LASSO (SR-LASSO)} problem
\be{
\label{SRLASSOproblem}
\min_{\bm{z} \in \cV^N_h} \lambda \nm{\bm{z}}_{\cV,1} + \nm{\bm{A} \bm{z} - \bm{b} }_{\cV,2},
}
where $\lambda > 0$ is a parameter. This is based on ideas of \cite{adcock2019correcting}. While the other approaches are more common, they have the undesirable feature that the optimal choices for their parameters (in the sense that they give the best error bounds) depend on the magnitude of the terms $\bm{e}$ and $\bm{e}'$. These terms are unknown (in particular, they typically depend on $f$). Hence, they would not give rise to a result such as Theorem \ref{t:mainthm1}, where the parameter choice is independent of $f$. Fortunately, as shown in \cite{adcock2019correcting}, the SR-LASSO \eqref{SRLASSOproblem} has such a desirable property.

The theory of Hilbert-valued compressed sensing has been developed in \cite{dexter2019mixed}. We recall several key definitions. The \textit{support} of a Hilbert-valued vector $\bm{x} = (x_i)^{N}_{i=1}\in \cV^N$ is
\bes{
\mathrm{supp}(\bm{x}) = \{ i : \nm{x_i}_{\cV} \neq 0 \} \subseteq \{1,\ldots,N\}.
}
A Hilbert-valued vector is \textit{$s$-sparse} if
\bes{
\nm{\bm{x}}_{\cV,0} : = | \mathrm{supp}(\bm{x}) | \leq s.
}
We write $\Sigma_{s} \subseteq \cV^N$ for the set of $s$-sparse Hilbert-valued vectors. For nonsparse vectors, we also define the $\ell^p$-norm \textit{best $s$-term approximation error} as
\bes{
\sigma_{s}(\bm{x})_{\cV,p} = \inf_{\bm{z} \in \cV^N} \{ \nm{\bm{x} - \bm{z}}_{\cV,p} : \bm{z} \in \Sigma_s \},\quad \bm{x} \in \cV^N.
}

We now focus on properties of the matrix $\bm{A}$ that ensure recovery of approximately sparse vectors via \eqref{SRLASSOproblem}. To this end, we now define some additional notation. Given a set $S \subseteq \{1,\ldots,N\}$ we write $P_{S} : \cV^N \rightarrow \cV^N$ for the projection that, given a vector $\bm{x} \in \cV^N$, produces a vector $P_{S}(\bm{x})$ whose $i$th entry is equal to $x_i$ if $i \in S$ and zero otherwise.

\begin{definition}
The matrix $\bm{A} \in \bbR^{m \times N}$ satisfies the \textit{robust Null Space Property (rNSP)} of order $1 \leq s \leq N$ over $\cV^N$ with constants $0 < \rho < 1$ and $\tau > 0$ if
\bes{
\nm{P_{S}(\x)}_{\cV,2} \leq  \dfrac{\rho \nm{P_{S^c} (\x)}_{\cV,1} }{\sqrt{s}}+\tau \nm{\bm{A} \x}_{\cV,2} ,\quad \forall \x \in \cV^N,
}
for any $S \subseteq [N]$ with $|S| \leq s$.
\end{definition}

See \cite[Defn.\ 4.1]{dexter2019mixed}. In \cite[Prop.\ 4.2]{dexter2019mixed} the authors show that the rNSP is sufficient to provide an error bound for minimizers of the quadratically-constrained basis pursuit problem. We need an analogous result for the SR-LASSO problem \eqref{SRLASSOproblem}. The following is a straightforward extension of \cite[Thm.\ 6.4]{adcock2021compressive}\footnote{Chapter available online at \url{www.compressiveimagingbook.com}.} from the scalar to the Hilbert-valued case.

\begin{lemma}\label{l:rNSP_SRLASSO}
Suppose that $\bm{A} \in \bbR^{m \times N}$ has the rNSP of order $1 \leq s \leq N$ with constants $0 < \rho < 1$ and $\tau > 0$. Let $\bm{x} \in \cV^N$, $\bm{b} = \bm{A} \bm{x} + \bm{e} \in \cV^m$ and
\bes{
\lambda \leq \frac{C_1}{C_2 \sqrt{s}},
}
where $C_1 = \frac{(3 \rho+1)(\rho+1)}{2(1-\rho)}$ and $C_2 = \frac{(3 \rho+5) \tau}{2(1-\rho)}$. Then every minimizer $\hat{\bm{x}} \in \cV^N$ of the Hilbert-valued SR-LASSO problem
\bes{
\min_{\bm{z} \in \cV^N} \lambda \nm{\bm{z}}_{\cV,1} + \nm{\bm{A} \bm{z} - \bm{b}}_{\cV,2},
}
satisfies
\bes{
\nm{\hat{\bm{x}} - \bm{x}}_{\cV,2} \leq 2 C_1 \frac{\sigma_{s}(\bm{x})_{\cV,1}}{\sqrt{s}} + \left ( \frac{C_1}{\sqrt{s} \lambda} + C_2 \right ) \nm{\bm{e}}_{\cV,2}.
}
\end{lemma}

Now we recall that a matrix $\bm{A} \in \bbR^{m \times N}$ satisfies the \textit{Restricted Isometry Property (RIP)} of order $1 \leq s \leq N$ with constant $0 < \delta_{s} < 1$ if $\delta_{s}$ is the smallest constant $\delta$ for which
\bes{
(1-\delta) \nm{\bm{x}}^2_2 \leq \nm{\bm{A} \bm{x}}^2_2 \leq (1+\delta) \nm{\bm{x}}^2_2,\quad \forall \mbox{$\bm{x} \in \bbR^N$, $\bm{x}$ $s$-sparse}.
}
The following result connects the RIP and rNSP. It is based on a combination of  \cite[Prop 4.3]{dexter2019mixed} (which applies to the Hilbert-valued case) and \cite[Lem.\ 5.17]{adcock2021compressive}
(which gives explicit values for $\rho$ and $\tau$):

\begin{lemma}
\label{l:RIPimpliesrNSP}
Suppose that $\bm{A} \in \bbR^{m \times N}$ satisfies the RIP of order $2s$ with constant $\delta_{2s} < \sqrt{2}-1$. Then $\bm{A}$ satisfies the rNSP of order $s$ over $\cV^N$ with constants $\rho$ and $\tau$ given by
\bes{
\rho = \frac{\sqrt{2} \delta_{2s}}{1-\delta_{2s}},\qquad \tau = \frac{\sqrt{1+\delta_{2s}}}{1-\delta_{2s}}.
}
\end{lemma}

With this in mind, we end this section with a result asserting that conditions under which the matrix defined in \eqref{def-measMatrix} satisfies the RIP:

\begin{lemma}
\label{l:LegMat_RIP}
Let $\{ \Psi_{\bm{\nu}} \}_{\bm{\nu} \in \bbN^d_0}$ be the orthonormal tensor Legendre polynomial basis of $L^2_{\varrho}(\cU)$ and $\bm{y}_1,\ldots,\bm{y}_m$ be drawn independently from the uniform measure on $\cU$. Let $0 < \delta, \varepsilon < 1$ and suppose that
\bes{
m \geq c \cdot 2^d \cdot s^2 \cdot \log(2s) \cdot \left ( \log(2 s) \cdot \min \{ \log(s) + d , \log(2d) \cdot \log(2s) \} + \log(\varepsilon^{-1}) \right ),
}
where $c>0$ is a universal constant.
Then, with probability at least $1-\varepsilon$, the matrix $\bm{A}$ defined in satisfies the RIP of order $s$ with constant $\delta_{s} \leq 1/4$.
\end{lemma}

Note that the choice of $1/4$ here is arbitrary. Any value less than $\sqrt{2}-1 \approx 0.41$ will suffice.

\begin{proof}
Theorem 2.2 of \cite{chkifa2018polynomial} implies that any matrix of this form satisfies the RIP of order $s$ with constant $\delta_{s} \leq 1/4$, provided (after simplifying)
\be{
\label{m_BOS_dexter_et_al}
m \gtrsim \Theta^2 \cdot s \cdot  \log(2 \Theta^2 s) \cdot \max \left \{ \log(2\Theta^2 s \log(2 \Theta^2 s)) \cdot \log(2N) , \log(2 \varepsilon^{-1} \log(2 \Theta^2 s) ) \right \},
}
where $N = |\Lambda|$ and $\Theta$ is defined by
\bes{
\Theta = \max_{\bm{\nu} \in \Lambda} \nm{\Psi_{\bm{\nu}}}_{L^{\infty}(\cU)}.
}
The Legendre polynomials attain their maxima at $\bm{y} = \bm{1}$ (see, for example, \cite{szego1975orthogonal}) and, due to the normalization with respect to the uniform measure, satisfy $| \Psi_{\bm{\nu}}(\bm{1}) | = \prod^{d}_{k=1} \sqrt{2 \nu_k + 1}$. Hence, since $\Lambda$ is the hyperbolic cross of index $s-1$ (recall \eqref{HC_index_set}), we have
\bes{
\Theta^2 \leq \max \left \{ \prod^{d}_{k=1} (2 \nu_k + 1) : \prod^{d}_{k=1} (\nu_k + 1) \leq s \right \} \leq 2^d s.
}
Furthermore, it was also show in \cite[Lem.\ 3.5]{chkifa2018polynomial} that $\Theta^2 \leq s^{\log(3) / \log(2)}$. Substituting this into the above expression and using the fact that
\bes{
\log(2 \Theta^2 s) \leq \log(2) + (\log(3)/\log(2) + 1) \log(s) \lesssim \log(2s) \leq 2 s.
}
now shows that \eqref{m_BOS_dexter_et_al} is implied by
\be{
\label{m_BOS_Leg_case}
m \gtrsim 2^d \cdot s^2 \cdot \log(2s) \cdot \left ( \log(2 s) \cdot \log(2N) + \log(\varepsilon^{-1}) \right ).
}
Furthermore, it can be shown that
\begin{equation}\label{Bound_N}
N \leq  \min \left \{ 2 s^3 4^d , \E^2 s^{2 + \log(d)/\log(2)} , \frac{s (\log(s) + d \log(2))^{d-1}}{(d-1)!}  \right \}.
\end{equation}
The first and third bounds are due to Theorems 3.7 and 3.5 of \cite{chernov2016new} respectively with values $s = d$, $a = 1$ and $T = s$ (note that there is a small typo in the statement of Theorem 3.5 of \cite{chernov2016new}: the denominator should read $\ln T - s \ln(a-1/2) + s - 1$ \cite{dung2020private}). The second bound is due to \cite[Thm.\ 4.9]{kuhn2015approximation} (note that, although \cite[Thm.\ 4.9]{kuhn2015approximation} has the condition $s \leq 2^d$ among its assumptions, an inspection of the proof reveals that this assumption is not necessary). In particular,
\eas{
\log(2N) &\leq \min \left \{ \log(8 s^3) + d \log(4) , (2 + \log(d)/\log(2)) \log(s) + 2 \right \}
\\
& \lesssim \min \{ \log(s) + d , \log(2d) \cdot \log(2 s) \}.
}
Substituting this into \eqref{m_BOS_Leg_case} now gives the result.
\end{proof}

\subsection{Polynomial approximation error bounds for holomorphic functions}\label{ss:polyapproxholobds}

In order to establish exponential rates of convergence, we need the following result regarding estimates for the error of the best $s$-term polynomial approximation of a holomorphic function $f$. For this next result, we recall that a set $S \subseteq \bbN^d_0$ is \textit{lower} if whenever $\bm{\nu} \in S$ then $\bm{\mu} \in S$ for all multi-indices $\bm{\mu}$ satisfying $\bm{\mu} \leq \bm{\nu}$.

\begin{theorem}
\label{t:best_s_term_poly}
Let $d \in \bbN$, $f : \cU \rightarrow \cV$ be holomorphic in a Bernstein polyellipse $\cE_{\bm{\rho}}$ for some $\bm{\rho} > \bm{1}$. Then for every $\epsilon >0$ there exists $\bar{s} = \bar{s}(d , \epsilon,\bm{\rho})$ such that, for every $s \geq \bar{s}$, there is a lower set $S \subset \bbN^d_0$ of size $|S| \leq s$ for which
\eas{
\nm{f - f_S}_{L^2_{\varrho}(\cU ; \cV)} \leq \nm{f - f_S}_{L^{\infty}(\cU;\cV)} & \leq \sum_{\bm{\nu} \notin S} \nm{\Psi_{\bm{\nu}} }_{L^{\infty}(\cU)}   \nm{ c_{\bm{\nu}} }_{\cV}
\\
& \leq \nm{f}_{L^{\infty}(\cE_{\bm{\rho}} ; \cV)} \exp \left ( -\frac{1}{d+1} \left ( \frac{s d! \prod^{d}_{j=1} \log(\rho_j) }{1+\epsilon} \right )^{1/d} \right ).
}
Furthermore, the same bound also applies to the coefficient error $\nm{\bm{c} - \bm{c}_{S} }_{1,\cV}$, where $\bm{c} = (c_{\bm{\nu}})_{\bm{\nu} \in \bbN^d_0}$ is the sequence of coefficients of $f$ as in \eqref{f_exp_Leg} and $\bm{c}_{S}$ is the infinite sequence with $\bm{\nu}$th entry equal to $c_{\bm{\nu}}$ if $\bm{\nu} \in S$ and zero otherwise.
\end{theorem}

Note that this result immediately implies \eqref{best_s_term_poly}. In fact, it is stronger, since it asserts an $L^{\infty}$-norm bound, and shows that this can be achieved by a set that is also lower.

\begin{proof}
The proof is mainly based on \cite[Theorem~3.5]{opschoor2019exponential}. We start by proving the theorem in the scalar-valued case, i.e.\ for $\cV = \bbR$. Note that in this case the coefficients $c_{\bm{\nu}}\in \bbR$. Inspecting the proof of \cite[Theorem~3.5]{opschoor2019exponential} we see that, for any given $\tau \in (0,1)$, choosing the multi-index set as
$$
S = S_{\tau} := \left\{\bm{\nu}\in\mathbb{N}_0^d : \bm{\rho}^{-\bm{\nu}} \geq \tau\right\},
$$
where $\bm{\rho}^{-\bm{\nu}} := \prod_{j= 1}^d \rho_j^{-\nu_j}$, leads to the upper bound
\begin{equation}
\label{eq:exp_bound_in_tau}
\sum_{\bm{\nu} \notin S_\tau} \nm{\Psi_{\bm{\nu}} }_{L^{\infty}(\cU)}   |c_{\bm{\nu}} |
\leq C \,  \|f\|_{L^\infty(\cE_{\bm{\rho}})} \, \tau \,  (1 + \log(1/\tau))^{2d},
\end{equation}
where $C = C(d,\bm{\rho}) >0$ is a constant depending only on $d$ and $\bm{\rho}$.
(Specifically, using the notation of \cite[Theorem~3.5]{opschoor2019exponential}, \eqref{eq:exp_bound_in_tau} is obtained by letting $\varepsilon = \tau$ and $k = 0$.)

The next step is to convert the upper bound \eqref{eq:exp_bound_in_tau} into an exponential best $s$-term decay rate. Let us consider the right-hand side of \eqref{eq:exp_bound_in_tau}. For any $\alpha \in (0,1)$, we have
$$
\lim_{\tau \to 0^+} \frac{\tau(1+\log(1/\tau))^{2d}}{\tau^\alpha} = 0.
$$
Therefore, for any $\alpha \in (0,1)$, we choose the largest $\bar{\tau} = \bar{\tau}(\alpha, d, \bm{\rho}) \in (0,1)$ such that
$$
\tau(1+\log(1/\tau))^{2d}
\leq \frac{\tau^{\alpha}}{C \exp\left(\alpha \sum_{j=1}^d \log(\rho_j)\right)},
\quad \forall 0 < \tau \leq \bar{\tau}.
$$
Combining the above inequality with \eqref{eq:exp_bound_in_tau} yields
\begin{equation}
\label{eq:exp_rates:sigma_s_leq_power_eps}
\sum_{\bm{\nu} \notin S_\tau} \nm{\Psi_{\bm{\nu}} }_{L^{\infty}(\cU)}   |c_{\bm{\nu}} |
\leq \frac{\tau^\alpha}{\exp\left(\alpha \sum_{j=1}^d \log(\rho_j)\right)}, \quad \forall 0 < \tau \leq \bar{\tau}.
\end{equation}
Now, following \cite[Theorem 5.2]{adcock2020gap}, we establish a direct link between the parameter $\tau \in (0,1)$ and the sparsity $s\in\bbN$. Indeed, for any $s \geq 2$ there exists only one value $\tau = \tau(s) \in (0,1)$ such that
\begin{equation}
\label{eq:s_tau_relation}
s = \prod_{j = 1}^d \left(\frac{\log(1/\tau)}{\log(\rho_j)}+1\right),
\end{equation}
and that
$$
\frac{s}{(d+1)^d} \leq |S_{\tau(s)}| \leq s.
$$
Observing that \eqref{eq:s_tau_relation} defines a monotone decreasing relation between $\tau(s)$ and $s$, we can find the minimum value $\bar{s} = \bar{s}(\alpha,d,\bm{\rho}) \in \mathbb{N}$ with $\bar{s} \geq 2$, such that for every $s \geq \bar{s}$, we have $\tau(s) \leq \bar{\tau}$. In this way, we have
$$
\sum_{\bm{\nu} \notin S_{\tau(s)}} \nm{\Psi_{\bm{\nu}} }_{L^{\infty}(\cU)}   |c_{\bm{\nu}} |
\leq \frac{\tau(s)^\alpha}{\exp\left(\alpha \sum_{j=1}^d \log(\rho_j)\right)}, \quad \forall s \geq \bar{s}.
$$
Now, we observe that the cardinality of $S_{\tau}$ can be explicitly bounded as in \cite[Lemma~3.3]{opschoor2019exponential} using a volumetric argument (note that $S_\tau$ corresponds to $\Lambda_{\varepsilon}$ in \cite{opschoor2019exponential}). This cardinality bound can be written as
$$
\tau
\leq
\exp\bigg(
\sum_{j = 1}^d \log(\rho_j) - \bigg(|S_\tau|d! \prod_{j = 1}^d \log(\rho_j)\bigg)^{\frac1d}\bigg), \quad \forall \tau \in (0,1).
$$
Combining the above inequalities and recalling that $|S_{\tau(s)}| \geq s/(d+1)^d$, we obtain that, for any $s \geq \bar{s}$,
\begin{align*}
\sum_{\bm{\nu} \notin S_{\tau(s)}} \nm{\Psi_{\bm{\nu}} }_{L^{\infty}(\cU)}   |c_{\bm{\nu}} |
& \leq
\exp\bigg(- \alpha\bigg(|S_{\tau(s)}|d! \prod_{j = 1}^d \log(\rho_j)\bigg)^{\frac{1}{d}}\bigg)  \\
& \leq \exp\bigg(-\frac{\alpha}{d+1}\bigg(sd! \prod_{j = 1}^d \log(\rho_j)\bigg)^{\frac{1}{d}}\bigg).
\end{align*}
Finally, we let $\alpha = 1/(1+\epsilon)^{1/d}$ and observe that $\nm{f - f_S}_{L^2_{\varrho}(\cU ; \cV)} \leq \nm{f - f_S}_{L^{\infty}(\cU;\cV)} \leq \sum_{\bm{\nu} \notin S} \nm{\Psi_{\bm{\nu}} }_{L^{\infty}(\cU)}  |c_{\bm{\nu}} |$. This concludes the proof in the case $\cV = \bbR$.

This proof can be generalized to the Hilbert-valued case by replacing coefficients' magnitude $|c_{\bm{\nu}}|$ with coefficients' norm $\|c_{\bm{\nu}}\|_{\cV}$. With this modification, the analogous of \eqref{eq:exp_bound_in_tau} holds. Indeed, the proof of \eqref{eq:exp_bound_in_tau} given in \cite[Theorem~3.5]{opschoor2019exponential} relies on coefficient bounds for Legendre polynomials that hold in the Hilbert-valued case as well (see, e.g., \cite[Lemma~3.15]{cohen2015high}. The rest of the argument is identical to the scalar-valued case. We also observe that the exponential bound holds for $\|\bm{c}-\bm{c}_S\|_{1,\cV}$ because $\|\Psi_{\bm{\nu}}\|_{L^\infty(\cU)} \geq 1$ for every $\bm{\nu} \in \bbN_0^d$. This completes the proof.
\end{proof}

\subsection{Proof of Theorem \ref{t:mainthm1}}

\label{ss:proof_main_thm}

We are now ready to prove the main result. Our strategy is based on reformulating the above compressed sensing formulation as a DNN training problem. We first require the following result \cite[Prop.\ 2.13]{opschoor2019exponential}:

\begin{proposition}
\label{Prop_Ex_NN}
There exists a universal constant $c> 0$ such that the following holds.
For every finite subset $\Lambda \subset \bbN_0^d$ and every $0< \delta <1$ there exists a ReLU neural network $\Phi_{\Lambda, \delta}: \bbR^{d} \rightarrow \bbR^{|\Lambda|}$ such that, if $\Phi_{\Lambda, \delta} =(\Phi_{\bnu, \delta})_{\bnu \in \Lambda}$, then
\begin{equation*}
\|\Psi_{\bnu} -\Phi_{\bnu, \delta} \|_{L^\infty(\cU)} \leq \delta,\quad \forall \bm{\nu} \in \Lambda.
\end{equation*}
The depth and size of this network satisfy
 \begin{equation*}
 \begin{split}
 \mathrm{depth}(\Phi)& \leq c \cdot (1+ d\log(d)) \cdot  (1+ \log(m(\Lambda))) \cdot (m(\Lambda)+\log(\delta^{-1})),
  \\
  \mathrm{size}(\Phi)& \leq c \cdot  (d^2m(\Lambda)^{2} +d m(\Lambda)\log(\delta^{-1})) +d^2 \cdot  |\Lambda| \cdot (1+ \log(m(\Lambda))+\log(\delta^{-1})),  \\
 \end{split}
 \end{equation*}
where $  m(\Lambda) = \max_{\bnu \in \Lambda} \|\bnu\|_{1}= \max_{\bnu \in \Lambda} \sum_{j=1}^d \nu_{j} $.
\end{proposition}

 The general idea of the proof is to use Proposition \ref{Prop_Ex_NN} to approximate the matrix-vector multiplication $\bA \z$ -- which is a polynomial evaluated at the sample points $\y_i$ -- as a neural network $\Phi$ evaluated at the sample points, or equivalently, to approximate the matrix $\bA$, which is built from polynomials, as a matrix $\bA'$ built from DNNs. We then use compressed sensing results applied to $\bA'$ to establish an error bound. Since this process commits an error, we first require the following result, which shows that the rNSP is robust to small matrix perturbations:

\begin{lemma}
\label{l:rNSPperturb}
Suppose that $\bm{A} \in \bbR^{m \times N}$ has the rNSP of order $1 \leq s \leq N$ over $\cV^N$ with constants $0 < \rho < 1$ and $\tau > 0$. Let $\bm{A}' \in \bbR^{m \times N}$ be such that $\nm{\bm{A} - \bm{A}'}_2 \leq \delta$, where
\bes{
0 \leq \delta < \frac{1-\rho}{\tau(\sqrt{s}+1)}.
}
Then $\bm{A}'$ has the rNSP of order $s$ over $\cV^N$ with constants $0 < \rho' < 1$ and $\tau' >0$ given by
\bes{
\rho' = \frac{\rho + \tau \delta \sqrt{s}}{1-\tau \delta},\quad \tau' = \frac{\tau}{1-\tau \delta}.
}
\end{lemma}
This is a straightforward extension of \cite[Lemma~8.5]{adcock2021compressive} from the scalar-valued case to the Hilbert-valued case. We give the proof for completeness:

\begin{proof} 
Let $\z \in \cV^N$ and a set $S \in [N]$ such that $|S| \leq s$.  Then, since $A$ has the rNSP,
\eas{
\nmu{P_{S} (\z)}_{\cV,2} &\leq \frac{\rho}{\sqrt{s}} \nmu{P_{S^c} (\z)}_{\cV,1} + \tau \nmu{\bm{A} \z}_{\cV,2}
\\
& \leq \frac{\rho}{\sqrt{s}} \nmu{P_{S^c} (\z)}_{\cV,1} + \tau \nmu{\bm{A}' \z}_{\cV,2} + \tau \delta \nmu{\z}_{\cV,2}
\\
& \leq \frac{\rho}{\sqrt{s}} \nmu{P_{S^c} (\z)}_{\cV,1} + \tau \nmu{\bm{A}' \z}_{\cV,2} + \tau \delta \nmu{P_{S} (\z)}_{\cV,2} + \tau \delta \nmu{P_{S^c} (\z)}_{\cV,1}.
}
Rearranging, we get
\bes{
(1 - \tau \delta ) \nmu{P_{S} (\z)}_{\cV,2} \leq  \frac{\rho + \tau \delta \sqrt{s}}{\sqrt{s}}  \nmu{P_{S^c} (\z)}_{\cV,1} + \tau \nmu{\bm{A}' \z}_{\cV,2}.
}
The result now follows immediately from the assumptions on $\delta$.
\end{proof}

\begin{proof} (of Theorem \ref{t:mainthm1}) We divide the proof into several steps:

\vspace{1pc} \noindent
\textit{Step 1:} We first define the class of DNNs $\cN$. Let $\widetilde{m}$ be as in \eqref{tildemdef} and define
\be{
\label{sdef}
s : = \left \lceil \left ( \widetilde{m}/2^d \right )^{1/2}  \right \rceil.
}
Now let $\Lambda = \Lambda^{\mathrm{HC}}_{s-1}$ and $\Phi_{\Lambda,\delta}$ be as in Proposition \ref{Prop_Ex_NN}. We will choose $\delta$ in Step 5. Letting $N = |\Lambda|$, we define the class $\cN$ as 
\bes{
\cN = \left \{ \Phi : \bbR^d \rightarrow \bbR^K : \Phi(\bm{y}) = \bm{Z}^{\top} \Phi_{\Lambda,\delta}(\bm{y}),\ \bm{Z} \in \bbR^{N \times K} \right \}.
}
Note that $\bm{Z}$ is the matrix of trainable parameters (it is the weight matrix in the output layer). We will establish the size, depth and number of trainable parameters of this class (part (a) of Theorem \ref{t:mainthm1}) in Step 6.

\vspace{1pc} \noindent
\textit{Step 2:} We next show that \eqref{trainingprob} can be reinterpreted as a SR-LASSO minimization problem. Analogously to \eqref{def-measMatrix}, we define the matrix $\bm{A}'$ as 
\bes{
\bm{A}' = \left(\frac{\Phi_{\bnu_j,\delta} (\y_i)}{\sqrt{m}} \right)^{m,N}_{i,j=1} \in \bbR^{m \times N},
}
where $\Phi_{\bm{\nu},\delta}$ is the $\bm{\nu}$th component of $\Phi_{\Lambda,\delta}$. Let $\Phi = \bm{Z}^{\top} \Phi_{\Lambda,\delta} \in \cN$ and $f_{\Phi}$ be as in \eqref{fPhi_DNN}. Then
\be{
\label{fPhi_in_c}
f_{\Phi}(\bm{y}_i) = \sum^{K}_{k=1} \left ( \Phi(\bm{y}_i) \right )_k\varphi_{k} = \sqrt{m} \sum^{K}_{k=1} (\bm{A}' \bm{Z})_{ik} \varphi_k = \sqrt{m} \left ( \bm{A}' \bm{z} \right )_i,
}
where $\bm{z} = (z_{\bm{\nu}_j})^{N}_{j=1} \in \cV^N_h$ is the Hilbert-valued vector with $\bm{\nu}_{j}$th component $z_{\bm{\nu}_j} = \sum^{K}_{k=1} Z_{jk} \varphi_k$. Hence
\bes{
\left ( \frac{1}{\sqrt{m}} f_{\Phi}(\bm{y}_i) \right )^{m}_{i=1} = \bm{A}' \bm{z}.
}
We now define the regularization function $\cJ$ as
\bes{
\cJ(\Phi) = \nm{\bm{z}}_{\cV,1},\footnote{Note that a DNN $\Phi$ is not, in general, uniquely defined by its parameters. In this case, this expression may fail to define a well-defined map $\cJ$.  However, this is not a problem since we can define $\cJ(\Phi)$ in this case as the infimum of $\nm{\bm{z}}_{\cV,1}$ over all parameters $\bm{Z}$ that yield the same DNN $\Phi$.}
}
We verify that $\cJ$ is equivalent to a norm over the trainable parameters $\bm{Z}$ (part (b) of Theorem \ref{t:mainthm1}). Notice that
\bes{
\nm{\bm{z}}_{\cV,1} = \sum^{N}_{j=1} \nm{z_{\bm{\nu}_{j}}}_{\cV} = \sum^{N}_{j=1} \nm{\sum^{K}_{k=1} Z_{jk} \varphi_k}_{\cV} = \sum^{N}_{j=1} \nm{\bm{Z}^{\top} \bm{e}_j }_{\bm{G},2} = \nm{\bm{G}^{1/2} \bm{Z}^{\top}}_{2,1}
}
where $\bm{e}_j \in \bbR^N$ is the $j$th coordinate vector, $\bm{G}^{1/2}$ is the unique positive definite square root of $\bm{G}$ (defined in \S \ref{s:mainres}, along with the norm $\|\cdot\|_{\bm{G},2})$
and $\nm{\cdot}_{2,1}$ is the matrix $\ell^{2,1}$-norm, defined in \S \ref{s:setup}.
Hence, $\cJ$ is equivalent to a norm over the trainable parameters.

Using this and \eqref{fPhi_in_c}, we now see that \eqref{trainingprob} can be expressed as the Hilbert-valued SR-LASSO problem
\be{
\label{SRLASSO_equiv_DNNtraining}
\min_{\bm{z} \in \cV^N_h} \lambda \nm{\bm{z}}_{\cV,1} + \nm{\bm{A}' \bm{z} - \bm{b}}_{\cV,2},
}
where $\bm{b} = \frac{1}{\sqrt{m}} (d_i)^{m}_{i=1} \in \cV^m_h$. We will choose the value of $\lambda$ in Step 4(ii). In particular, any minimizer $\bm{\hat{c}} = (\hat{c}_{\bm{\nu}} )_{\bm{\nu} \in \Lambda}$ of \eqref{SRLASSO_equiv_DNNtraining} yields a minimizer $\hat{\Phi}$ of \eqref{trainingprob}, given by
\bes{
\hat{\Phi} =\widehat{\bm{C}}^{\top} \Phi_{\Lambda,\delta}(\bm{y}),
}
where $\widehat{\bm{C}} = \left ( \widehat{C}_{jk} \right )^{N,K}_{j,k=1}$ is such that
\bes{
\hat{c}_{\bm{\nu}_j} = \sum^{K}_{k=1} \widehat{C}_{jk} \varphi_k,\quad \forall j = 1,\ldots,N,
}
and vice versa. This completes Step 2.

\vspace{1pc} \noindent
\textit{Step 3:} In the third step, we show that the matrix $\bm{A}'$ has the rNSP of order $s$ over $\cV^N_h$ with probability at least $1-\varepsilon$.
Since $\widetilde{m} \geq 2^d \bar{s}^2 \geq 2^d$ by assumption, we have
\be{
\label{smtildeub}
s \leq 2  (\widetilde{m}/2^d)^{1/2},
}
and therefore $s \leq 2 (m/(2^d \cL))^{1/2} \leq m$, since $\cL \geq 1$ for a suitable choice of the universal constant $c_0$.
Hence, recalling \eqref{eq:def_log_factor} and \eqref{tildemdef}, we see that
\eas{
m = \widetilde{m} \cdot \cL(m,d,\varepsilon) \geq \widetilde{m} \cdot \cL(s,d,\varepsilon) \geq 2^d \cdot (s^2 /4)\cdot \cL(s,d,\varepsilon),
}
and therefore
\bes{
m \gtrsim 2^d \cdot s^2 \cdot \log(2s) \cdot \left (  \log(2s) \cdot \min \{ \log(2s)+d , \log(2s) \cdot \log(2d) \}+ \log(\varepsilon^{-1})  \right).
}
Lemma~\ref{l:LegMat_RIP} (after replacing $s$ by $2s$) now implies that the matrix $\bm{A}$ satisfies the RIP of order $2s$ with constant $\delta_{2s} \leq 1/4$ and Lemma \ref{l:RIPimpliesrNSP} implies that $\bm{A}$ satisfies the rNSP of order $s$ over $\cV^N_h$ with constants
\be{
\label{rho_tau_A_vals}
\rho = \sqrt{2}/3,\quad \tau = 2 \sqrt{5}/3,
}
with probability at least $1-\varepsilon$. To show that $\bm{A}'$ satisfies the rNSP we use Lemma \ref{l:rNSPperturb}. We first bound $\nm{\bm{A} - \bm{A}'}_2$. Let $\bm{z} \in \bbR^N$. Then we observe that
\eas{
\nm{(\bm{A} - \bm{A}') \bm{z}}^2_2 &= \frac1m \sum^{m}_{i=1} \left | \sum_{\bm{\nu} \in \Lambda} \left ( \Psi_{\bm{\nu}}(\bm{y}_i) - \Phi_{\bm{\nu},\delta}(\bm{y}_i) \right ) z_{\bm{\nu}} \right |^2
\\
& \leq \left ( \sum_{\bm{\nu} \in \Lambda} \nm{ \Psi_{\bm{\nu}}(\bm{y}_i) - \Phi_{\bm{\nu},\delta}(\bm{y}_i) }_{L^{\infty}(\cU)} | z_{\bm{\nu}} | \right )^2
\\
& \leq  \sum_{\bm{\nu} \in \Lambda}  \nm{ \Psi_{\bm{\nu}} - \Phi_{\bm{\nu},\delta} }^2_{L^{\infty}(\cU)} \nm{\bm{z}}^2_2 \leq N \delta^2 \nm{\bm{z}}^2_2,
}
by definition of $\Phi_{\bm{\nu},\delta}$.
Hence
\bes{
\nm{\bm{A} - \bm{A}'}_{2} \leq \sqrt{N} \delta.
}
Now suppose that $\delta$ satisfies
\be{
\label{delta_cond_1}
\sqrt{N} \delta \leq \frac{9 - 4 \sqrt{2}}{2 \sqrt{5}(3 + 4\sqrt{s})},
}
(the choice of $\delta$ that will be made in Step 5 will ensure this condition).
Using Lemma \ref{l:rNSPperturb} and the values \eqref{rho_tau_A_vals} for $\rho$ and $\tau$ we now see that $\bm{A}'$ has the rNSP of order $s$ over $\cV^N_h$ with constants
\be{
\label{rho_tau_prime}
\rho' = \frac34,\qquad \tau' = \frac{\sqrt{5}(3+4 \sqrt{s})}{2(\sqrt{2}+3 \sqrt{s})} \leq \frac{3\sqrt{5}}{2},
}
with probability at least $1-\varepsilon$.

\vspace{1pc} \noindent
\textit{Step 4:} We now derive an error bound for $f - f_{\hat{\Phi}}$, where $\hat{\Phi}$ is a minimizer of \eqref{trainingprob} and $f_{\hat{\Phi}}$ is as in \eqref{fPhi_DNN}. Throughout, we assume the results of the previous steps. In particular, the matrix $\bm{A}'$ defined in step 2 has the rNSP of order $s$ (step 3). The problem \eqref{SRLASSO_equiv_DNNtraining}, or equivalently \eqref{trainingprob}, involves three discretizations. Truncation of the infinite expansion via the index set $\Lambda$, discretization of the space $\cV$ via $\cV_h$ and replacement of the polynomial functions by DNNs. First, given $\hat{\Phi} = \widehat{\bm{C}}^{\top} \Phi_{\Lambda,\delta}$ we let $\hat{\bm{c}}$ be the corresponding minimizer   of \eqref{SRLASSO_equiv_DNNtraining} and set
\bes{
f_{\hat{\Psi}} = \sum_{\bm{\nu} \in \Lambda} \hat{c}_{\bm{\nu}} \Psi_{\bm{\nu}}.
}
We now proceed as follows:
\eas{
\nm{f -f_{\hat{\Phi}}}_{L^2_{\varrho}(\cU ; \cV)} 
\leq  & \nm{f - \cP_h(f)}_{L^2_{\varrho}(\cU; \cV)} + \nm{\cP_h(f) - \cP_{h}(f_{\Lambda}) }_{L^2_{\varrho}(\cU ; \cV)}
\\
& + \nm{\cP_h(f_{\Lambda}) - f_{\hat{\Psi}}}_{L^2_{\varrho}(\cU ; \cV)} + \nm{f_{\hat{\Psi}} - f_{\hat{\Phi}} }_{L^2_{\varrho}(\cU ; \cV)}
\\
 =: & \nm{f - \cP_h(f)}_{L^2_{\varrho}(\cU; \cV)} + A_1 + A_2 + A_3.
}
Here $\cP_h$ is as defined in \S \ref{ss:setup} and $f_{\Lambda}$ is as in \eqref{f_exp_trunc}. We now bound $A_1$, $A_2$ and $A_3$.

\vspace{1pc} \noindent \textit{Step 4(i):} We commence with $A_1$. This is elementary. Since $\cP_h$ is an orthogonal projection we have $\nm{\cP_h(a)}_{\cV} \leq \nm{a}_{\cV}$ for $a \in \cV$ and therefore
\be{
\label{A1bound}
A_1 
= \nm{\cP_h(f) - \cP_{h}(f_{\Lambda}) }_{L^2_{\varrho}(\cU ; \cV)} 
\leq \nm{f - f_{\Lambda}}_{L^2_{\varrho}(\cU ; \cV)} 
\leq \nm{f - f_{\Lambda}}_{L^{\infty}(\cU ; \cV)},
}
recalling that we are considering a probability measure $\D \varrho(\bm{y})$ over $\mathcal{U}$.

\vspace{1pc} \noindent \textit{Step 4(ii):} We next consider $A_2$.
Observe that
\bes{
\cP_{h}(f_{\Lambda}) - f_{\hat{\Psi}} = \cP_{h} \left ( \sum_{\bm{\nu} \in \Lambda} c_{\bm{\nu}} \Psi_{\bm{\nu}}  \right ) - \sum_{\bm{\nu} \in \Lambda} \hat{c}_{\bm{\nu}} \Psi_{\bm{\nu}} = \sum_{\bm{\nu} \in \Lambda} \left ( \cP_h(c_{\bm{\nu}}) - \hat{c}_{\bm{\nu}} \right ) \Psi_{\bm{\nu}}.
}
Define the vector $\cP_{h}(\bm{c}_{\Lambda}) = (\cP_{h} (c_{\bm{\nu}}) )_{\bm{\nu} \in \Lambda}$. Then, by orthonormality of the $\Psi_{\bm{\nu}}$'s, we have
\bes{
\nm{\cP_{h}(f_{\Lambda}) - f_{\hat{\Psi}} }_{L^2_{\varrho}(\cU ; \cV)} = \nm{\cP_{h}(\bm{c}_{\Lambda}) - \hat{\bm{c}} }_{\cV,2}.
}
To bound this term, apply Lemma \ref{l:rNSP_SRLASSO} to the problem \eqref{SRLASSO_equiv_DNNtraining}. This gives
\bes{
\nm{\cP_h(\bm{c}_{\Lambda}) - \hat{\bm{c}} }_{\cV,2} \leq 2 C'_1 \frac{\sigma_s(\cP_h(\bm{c}_{\Lambda}))_{\cV,1}}{\sqrt{s}} + \left ( \frac{C'_1}{\sqrt{s} \lambda} + C'_2 \right ) \nm{\bm{A}' \cP_h(\bm{c}_{\Lambda}) - \bm{b} }_{\cV,2},
}
where  $C'_1 = \frac{(3 \rho'+1)(\rho'+1)}{2(1-\rho')}$ and $C_2 = \frac{(3 \rho'+5) \tau'}{2(1-\rho')}$ with $\rho'$ and $\tau'$ as in \eqref{rho_tau_prime}. This holds provided $\lambda \leq C'_1 / (C'_2 \sqrt{s})$. Thus, we now set
\be{
\label{lambda_def}
\lambda = \frac{C'_1}{C'_2 \sqrt{s}}.
}
Since $s$ is given via \eqref{sdef} in terms of $\widetilde{m}$ and $d$, we have shown part (c) of Theorem \ref{t:mainthm1}. Notice that $C'_1 , C'_2  \asymp 1$ due to the values of $\rho'$ and $\tau'$ given by \eqref{rho_tau_prime}. Hence
\bes{
\nm{\cP_h(\bm{c}_{\Lambda}) - \hat{\bm{c}} }_{\cV,2} \lesssim \frac{\sigma_s(\cP_h(\bm{c}_{\Lambda}))_{\cV,1}}{\sqrt{s}} + \nm{\bm{A}' \cP_h(\bm{c}_{\Lambda}) - \bm{b} }_{\cV,2}.
}
Consider the first term. Let $\pi : \{1,\ldots,N\} \rightarrow \Lambda$ be a bijection that gives a nonincreasing rearrangement of the sequence $(\nm{c_{\bm{\nu}}}_{\cV})_{\bm{\nu} \in \Lambda}$. Then
\bes{
\sigma_{s}(\cP_h(\bm{c}_{\Lambda}))_{\cV,1} \leq \sum^{N}_{i=s+1} \nm{\cP_h(c_{\bm{\nu}_{\pi(i)}})}_{\cV} \leq \sum^{N}_{i=s+1} \nm{c_{\bm{\nu}_{\pi(i)}}}_{\cV} = \sigma_{s}(\bm{c}_{\Lambda})_{\cV,1} \leq \sigma_{s}(\bm{c})_{\cV,1}.
}
Hence,
\be{
\label{SRLASSO_gives_this}
\nm{\cP_h(\bm{c}_{\Lambda}) - \hat{\bm{c}} }_{\cV,2} \lesssim \frac{\sigma_s(\bm{c})_{\cV,1}}{\sqrt{s}} + \nm{\bm{A}' \cP_h(\bm{c}_{\Lambda}) - \bm{b} }_{\cV,2}.
}
We now estimate the second term. Let $i = 1,\ldots,m$ and write
\eas{
\sqrt{m} \left ( \bm{A}' \cP_{h}(\bm{c}_{\Lambda}) - \bm{b} \right )_i &= \sum_{\bm{\nu} \in \Lambda} \cP_h(c_{\bm{\nu}}) \Phi_{\bm{\nu} , \delta}(\bm{y}_i) - f(\bm{y}_i) - n_i
\\
& = \sum_{\bm{\nu} \in \Lambda} \cP_h(c_{\bm{\nu}}) \left ( \Phi_{\bm{\nu} , \delta}(\bm{y}_i) - \Psi_{\bm{\nu}}(\bm{y}_i) \right ) + \cP_h(f_{\Lambda})(\bm{y}_i) - f(\bm{y}_i) - n_i.
}
Then 
\eas{
\|  \sqrt{m}  ( \bm{A}' \cP_{h} & (\bm{c}_{\Lambda}) - \bm{b}  )_i  \|_{\cV} 
\\
 & \leq \sum_{\bm{\nu} \in \Lambda} \nm{\cP_h(c_{\bm{\nu}}) }_{\cV} \delta + \nm{\cP_h(f_{\Lambda})(\bm{y}_i) -  f(\bm{y}_i)}_{\cV} + \nm{n_i}_{\cV}
\\
& \leq \delta \sum_{\bm{\nu} \in \Lambda} \nm{c_{\bm{\nu}}}_{\cV} + \nm{\cP_h(f_{\Lambda})(\bm{y}_i) - \cP_h(f)(\bm{y}_i) }_{\cV} + \nm{\cP_h(f)(\bm{y}_i) - f(\bm{y}_i) }_{\cV} + \nm{n_i}_{\cV}
\\
& \leq \sqrt{N} \delta \nm{\bm{c}_{\Lambda}}_{\cV,2} + \nm{f - f_{\Lambda}}_{L^{\infty}(\cU ; \cV)} + \nm{f - \cP_h(f)}_{L^{\infty}(\cU ; \cV)} + \nm{n_i}_{\cV}.
}
Observing that $\nm{\bm{c}_{\Lambda}}_{\cV,2} \leq \nm{\bm{c}}_{\cV,2} = \nm{f}_{L^2_{\varrho}(\cU ; \cV)} \leq \nm{f}_{L^{\infty}(\cU ; \cV)} \leq 1$ by Parseval's identity and the assumption $f \in \cH\cA(\gamma,\epsilon,d)$, we conclude that
\bes{
\nm{\bm{A}' \cP_h(\bm{c}_{\Lambda}) - \bm{b} }_{\cV,2} \leq \sqrt{N} \delta + \nm{f - f_{\Lambda}}_{L^{\infty}(\cU ; \cV)} + \nm{f - \cP_h(f)}_{L^{\infty}(\cU ; \cV)} + \nm{\bm{e}}_{\cV,2},
}
where we recall from \eqref{E123_def} that $\bm{e}$ is defined by $\bm{e} = \frac{1}{\sqrt{m}} (n_i)^{m}_{i=1}$. Using the definitions of $E_2$ and $E_3$ from \eqref{E123_def} and substituting this into \eqref{SRLASSO_gives_this} now yields
\be{
\label{A2bound}
A_2 = \nm{\cP_h(\bm{c}_{\Lambda}) - \hat{\bm{c}} }_{\cV,2} \lesssim \frac{\sigma_s(\bm{c})_{\cV,1}}{\sqrt{s}} + \sqrt{N} \delta + \nm{f - f_{\Lambda}}_{L^{\infty}(\cU ; \cV)} +E_2 + E_3.
}

\vspace{1pc} \noindent \textit{Step 4(iii):} Finally, we consider $A_3$. We have
\bes{
\nm{f_{\hat{\Psi}} - f_{\hat{\Phi}}}_{L^2_{\varrho}(\cU ; \cV)} \leq \sum_{\bm{\nu} \in \Lambda} \nm{\Psi_{\bm{\nu}} - \Phi_{\bm{\nu},\delta} }_{L^2_{\varrho}(\cU)} \nm{\hat{c}_{\bm{\nu}}}_{\cV} \leq \delta \nm{\hat{\bm{c}}}_{\cV,1}.
}
We now use the fact that $\hat{\bm{c}}$ is a minimizer of \eqref{SRLASSO_equiv_DNNtraining} to get
\bes{
\lambda \nm{\hat{\bm{c}}}_{\cV,1} \leq \lambda \nm{\bm{0}}_{\cV,1} + \nm{\bm{A}' \bm{0} - \bm{b} }_{\cV,2} = \nm{\bm{b}}_{\cV,2}.
}
where $\bm{0} \in \cV^N_h$ is the zero vector. Using the definition of $\bm{b}$ (given after equation \eqref{SRLASSO_equiv_DNNtraining}) and \eqref{lambda_def}, we deduce that
\bes{
\nm{\hat{\bm{c}}}_{\cV,1} \lesssim \sqrt{s} \left ( \nm{\bm{e}}_{\cV,2}  + \nm{f}_{L^{\infty}(\cU ; \cV) } \right ) \leq \sqrt{s} \left ( \nm{\bm{e}}_{\cV,2} + 1 \right ) = \sqrt{s}(E_2+1).
}
Note that \eqref{delta_cond_1} implies that $\delta \sqrt{s} \lesssim 1/\sqrt{N} \lesssim 1$. 
Hence, we conclude that
\be{
\label{A3bound}
A_3 = \nm{f_{\hat{\Psi}} - f_{\hat{\Phi}}}_{L^2_{\varrho}(\cU ; \cV)} \lesssim \sqrt{s} \delta  +E_2.
}

\vspace{1pc} \noindent Combining the estimates \eqref{A1bound}, \eqref{A2bound} and \eqref{A3bound} and noticing that $s \leq N$ by definition, we deduce that
\be{
\label{everything_but_the_poly_bounds}
\nm{f -f_{\hat{\Phi}}}_{L^2_{\varrho}(\cU ; \cV)} 
\lesssim \nm{f - f_{\Lambda}}_{L^{\infty}(\cU ; \cV)} + \frac{\sigma_s(\bm{c})_{\cV,1}}{\sqrt{s}} + \sqrt{N} \delta +E_2 + E_3 .
}
This concludes Step 4.

\vspace{1pc}\noindent
\textit{Step 5:} In this penultimate step we use Theorem \ref{t:best_s_term_poly} to estimate the remaining error terms in \eqref{everything_but_the_poly_bounds} and give the exact value of $\delta$. Let $S$ be the set defined in Theorem \ref{t:best_s_term_poly}. Since $|S| \leq s$ and $\nm{\Psi_{\bm{\nu}}}_{L^{\infty}(\cU)} \geq 1$ (this is due to the fact the $\Psi_{\bm{\nu}}$ have unit $L^2_{\varrho}$-norm with respect to the uniform probability measure over $\cU$) we have
\bes{
\sigma_{s}(\bm{c})_{\cV,1} \leq \sum_{\bm{\nu} \notin S} \nm{c_{\bm{\nu}}}_{\cV} \leq \sum_{\bm{\nu} \notin S} \nm{\Psi_{\bm{\nu}} }_{L^{\infty}(\cU)}   \nm{ c_{\bm{\nu}} }_{\cV} .
}
Further, since $S$ is lower and $\Lambda = \Lambda^{\mathrm{HC}}_{s-1}$ we also have $S \subseteq \Lambda$. In fact, $\Lambda^{\mathrm{HC}}_{s-1}$ is the union of all lower sets of size at most $s$. It follows that
\bes{
\nm{f - f_{\Lambda}}_{L^{\infty}(\cU ; \cV)} \leq \sum_{\bm{\nu} \notin \Lambda} \nm{\Psi_{\bm{\nu}} }_{L^{\infty}(\cU)}   \nm{ c_{\bm{\nu}} }_{\cV} \leq \sum_{\bm{\nu} \notin S} \nm{\Psi_{\bm{\nu}} }_{L^{\infty}(\cU)}   \nm{ c_{\bm{\nu}} }_{\cV} .
}
Recall that $\widetilde{m} \geq 2^{d} \bar{s}^2$ by assumption. Hence \eqref{sdef} and \eqref{smtildeub} implies that $s \geq \bar{s}$.
We now use Theorem \ref{t:best_s_term_poly} and the fact that $f \in \mathcal{HA}(\gamma,\epsilon,d)$ to deduce that
\bes{
\sum_{\bm{\nu} \notin S} \nm{\Psi_{\bm{\nu}} }_{L^{\infty}(\cU)}   \nm{ c_{\bm{\nu}} }_{\cV} \leq \exp(-\gamma s^{1/d}).
}
Returning to \eqref{everything_but_the_poly_bounds}, this therefore gives
\bes{
\nm{f -f_{\hat{\Phi}}}_{L^2_{\varrho}(\cU ; \cV)} \lesssim \exp(-\gamma s^{1/d} ) + \sqrt{N} \delta + E_2 + E_3 \leq \exp(-\gamma s^{1/d}) + E_2 + E_3,
}
provided
\be{
\label{delta_cond_2}
\delta \leq N^{-1/2} \exp(-\gamma s^{1/d}).
}
Hence, recalling \eqref{delta_cond_1} we now set 
\be{
\label{delta_def}
\delta = \frac{1}{\sqrt{N}} \min \left \{ \frac{9 - 4 \sqrt{2}}{2 \sqrt{5}(3 + 4\sqrt{s})} , \exp(-\gamma s^{1/d}) \right \},
}
and recall the definitions \eqref{sdef} and \eqref{E123_def} of $s$ and $E_1$ respectively to deduce that
\bes{
\nm{f -f_{\hat{\Phi}}}_{L^2_{\varrho}(\cU ; \cV)} \lesssim E_1+E_2+E_3,
}
as required.

\vspace{1pc} \noindent \textit{Step 6:} Having now shown the error bound, the final step of the proof involves bounding the size, depth and number of trainable parameters for DNNs in the class $\cN$ defined in Step 1. To do this,  we notice from \eqref{delta_def} that
\[
\delta \gtrsim  s^{-1/2} N^{-1/2}\exp(-\gamma s^{1/d}),
\]
which, after a few steps, implies
\[
\log(\delta^{-1}) \lesssim  \log(s)+ \log(N) + \gamma s^{1/d}.
\]
Moreover, from the definition of the hyperbolic cross index set \eqref{HC_index_set} we notice that $m(\Lambda) \leq s$ (recall that $m(\Lambda) = \max_{\nu\in\Lambda}\|\bm{\nu}\|_1$, as defined in Proposition~\ref{Prop_Ex_NN}). Then, substituting this into  the bounds of the size and depth in  Proposition~\ref{Prop_Ex_NN}, we deduce that
 \begin{equation*}
 \begin{split}
 \mathrm{depth}(\cN)&  \lesssim    (1+ d\log(d)) \cdot  (1+ \log(s)) \cdot (s+\log(s)+ \log(N) + \gamma s^{1/d} ),
  \\
  \mathrm{size}(\cN)&  \lesssim    d^2s^2 +d s(\log(s)+ \log(N) + \gamma s^{1/d}) +d^2 \cdot  N \cdot (1+ \log(s)+ \log(N) + \gamma s^{1/d})+NK.  \\
 \end{split}
 \end{equation*}
We recall that $s \geq 1$. Then after some rearrangements we have
 \begin{equation}\label{size_depth}
 \begin{split}
 \mathrm{depth}(\cN)&  \lesssim    (1+ d\log(d)) \cdot  (1+ \log(s)) \cdot (s+ \log(N) + \gamma s^{1/d} ),
  \\
  \mathrm{size}(\cN)&  \lesssim    d^2s^2 + (ds+ d^2N)( \log(s)+ \log(N) + \gamma s^{1/d})+NK,  \\
 \end{split}
 \end{equation}
Now, it is easy to see from the definition in \eqref{sdef} that  \eqref{size_depth} becomes
 \begin{equation*}
 \begin{split}
 \mathrm{depth}(\cN)&  \lesssim    (1+ d\log(d)) \cdot  (1+ \log(\widetilde{m})) \cdot \left ( (\widetilde{m}/2^d )^{1/2}+ \log(N) + \gamma \widetilde{m}^{1/(2d)} \right ),
  \\
  \mathrm{size}(\cN)&  \lesssim    d^2(\widetilde{m}/2^d)  +    (d (\widetilde{m}/2^d)^{1/2}+ d^2N) \Big( \log(\widetilde{m})+ \log(N) + \gamma \widetilde{m}^{1/(2d)}\Big)+NK,  \\
 \end{split}
 \end{equation*}
Next we recall that the number of trainable parameters is $\mathrm{param}(\cN) =  N \cdot K$. To complete the proof, we use \eqref{Bound_N} and \eqref{smtildeub} to obtain $N \leq \Delta$, as required.
\end{proof}

\acks{BA acknowledges the support of the PIMS CRG ``High-dimensional Data
Analysis'', SFU's Big Data Initiative ``Next Big Question" Fund and NSERC
through grant R611675. SB acknowledges the support of NSERC, the Faculty of Arts and Science of Concordia University, and the CRM Applied Math Lab. ND acknowledges the support of the PIMS Postdoctoral Fellowship program.}

\end{document}